%% file: main.tex
\documentclass{article}

\PassOptionsToPackage{numbers, compress}{natbib}

\usepackage[preprint]{neurips_2026}

\usepackage[utf8]{inputenc} 
\usepackage[T1]{fontenc}    
\usepackage{hyperref}       
\usepackage{url}            
\usepackage{booktabs}       
\usepackage{amsfonts}       
\usepackage{nicefrac}       
\usepackage{microtype}      
\usepackage{xcolor}         

\usepackage{pifont}
\newcommand{\cmark}{\ding{51}}
\newcommand{\xmark}{\ding{55}}
\usepackage{xcolor} 
\usepackage{colortbl}
\usepackage{graphicx}
\usepackage{wrapfig}
\usepackage{amsmath}
\usepackage{amssymb}
\usepackage{mathtools}
\usepackage{amsthm}
\usepackage{float}  
\usepackage{makecell}
\usepackage{multirow}
\theoremstyle{plain}
\newtheorem{theorem}{Theorem}[section]
\newtheorem{proposition}[theorem]{Proposition}

\theoremstyle{definition}
\newtheorem{definition}[theorem]{Definition}

\theoremstyle{remark}

\title{SCAN: Enhance Time Series Anomaly Detection via Multi-Scale Neighborhood-Centered Clustering}

\author{Xingze Zheng$^{1}$, Hanyin Cheng$^{1}$, Siyuan Wang$^{1}$, Yiting Hao$^{1}$, \\ \textbf{Peng Chen$^{2}$, Yuan Jun$^{3}$, Yang Shu$^{1}\thanks{Corresponding author}$}\\
$^1$East China Normal University $^2$2012 APPLab, Huawei $^3$Huawei\\
\texttt{\{xzzheng,hycheng,sywang,ythao\}@stu.ecnu.edu.cn}, \\
\texttt{\{chenpeng192,yuanjun25\}@huawei.com},
\texttt{\{yshu\}@dase.ecnu.edu.cn} \\
}

\begin{document}

\maketitle

\begin{abstract}
Time series anomaly detection plays a crucial role in a wide range of real-world applications. Reconstruction-based methods have become the mainstream paradigm, but they suffer from over-generalization and under-generalization problems, which are challenging to balance. To address this, we introduce multi-scale clustering to enhance reconstruction-based methods. At the representation level, we integrate the cluster center representations of normal patterns to constrain the model to target representative normal patterns for reconstruction, preventing dominance of powerful capacity and representation capability. At the anomaly criterion level, we derive anomaly confidence score based on cluster membership probability and combine it with reconstruction error, providing dual criteria for detection. Furthermore, the effectiveness of the cluster center representations and anomaly confidence score depends on the clustering performance. Accordingly, we extract neighborhood-centered representations for multi-view clustering to improve clustering performance. Extensive experiments on multiple real-world datasets from diverse application domains demonstrate the state-of-the-art performance of SCAN.
\end{abstract}

\input{Sections/Introduction}

\input{Sections/Related_Work}

\input{Sections/Methodology}

\input{Sections/Experiments}
\input{Sections/Conclusion}






{
\small
\bibliographystyle{ieeetr}
\bibliography{main}
}


\newpage
\appendix
\input{Sections/Appendix}

\end{document}

%% file: Sections/Introduction.tex
\section{Introduction}
\label{Introduction}

With the rapid advancement of digitalization, time series analysis has been widely applied across diverse sectors, including finance, transportation, healthcare, meteorology, energy, and the Internet of Things~\cite{zhou2022fedformer,chen2024pathformer,wang2025lightgts}. Mining historical time
series data reveals intrinsic temporal patterns and latent regularities, providing reliable decision support for downstream tasks and practical decision-making~\cite{wu2021autoformer,Yuqietal-2023-PatchTST,chen2025aimts}. As a key branch, time series anomaly detection identifies outlier points or anomalous subsequences that deviate from normal patterns~\cite{wu2025catch}, enabling early detection of potential anomalies and timely implementation of preventive measures to mitigate risks and avoid heavy losses. 
Time series anomaly detection is categorized into supervised and unsupervised anomaly detection based on labeled data requirements~\cite{wang2024deep,qiu2025tab}. In practice, anomalous data is scarce and diverse, rendering its collection and labeling highly challenging. Thus, unsupervised anomaly detection has become a widely studied and applied practical solution~\cite{shentu2025towards}, which identifies anomalies solely by learning from normal data. Among these methods, reconstruction-based methods become the mainstream paradigm~\cite{li2025crossad}.
However, reconstruction-based methods suffer from over-generalization and under-generalization issues, as shown in Figure~\ref{fig:introduction_fig1}. Over-generalized model reconstructs simple anomalies as well as normal ones, causing high false negative rates, while under-generalized one fails to reconstruct complex normal patterns, leading to high false positive rates. In both cases, the reconstruction errors of normal and anomalous data lose their distinctiveness, making it impossible to detect anomalies. To mitigate over-generalization and under-generalization issues in reconstruction-based methods, two key research challenges remain:

\begin{figure}[h]  
    \centerline{\includegraphics[width=0.55\linewidth]{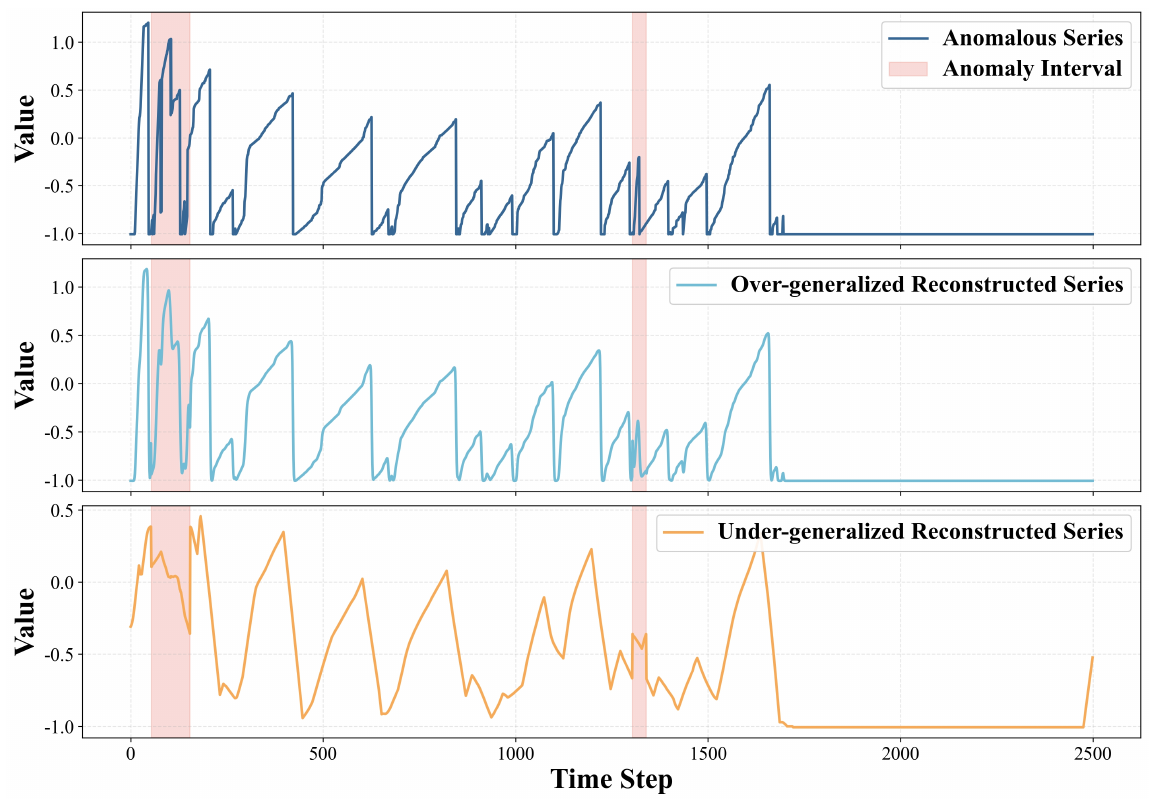}}
    \caption{Example of over-generalization and under-generalization. From top to bottom: anomalous series, over-generalized reconstructed series, and under-generalized one. Over-generalization reconstructs both normal and anomalous patterns, while under-generalization fails to reconstruct complex normal patterns.}
    \label{fig:introduction_fig1}
\end{figure}

First, \textbf{traditional reconstruction paradigms fail to balance over-generalization and under-generalization}. Learning normal sample representations and detecting anomalies via reconstruction error inherently involves a trade-off: powerful capacity and representation capability causes over-generalization~\cite{yoon2025momemto}, while inadequate representation learning leads to under-generalization. A single reconstruction dimension cannot concurrently balance suppressing simple anomaly miss-detection by narrowing the generalization boundary and reducing complex normal sample false detection by enhancing representation capability, thus limiting detection accuracy in complex temporal scenarios.
\textbf{To address this challenge}, we innovatively propose a novel anomaly detection paradigm by integrating clustering into reconstruction-based methods at the representation and anomaly criterion levels.
This is due to clustering can alleviate the above contradiction from the sample distribution perspective~\cite{qiu2025duet}. On one hand, it can capture representative normal patterns to constrain the scope of the model’s reconstruction objectives and thus suppress over-generalization. On the other hand, its cluster membership probabilities serve as another criterion for anomaly detection, thereby avoiding complete reliance on reconstruction-based results. 
Specifically, at the representation level, we first capture rich normal patterns via multi-scale modeling since single-scale modeling captures only limited normal patterns~\cite{li2025crossad}. Then, each scale generates cluster-weighted representations for its time series through weighted aggregation of all cluster center representations, followed by inter-scale fusion of these cluster-weighted representations and intra-scale fusion of the latter with original representations. This constrains the model to target representative normal patterns for reconstruction, thus suppressing over-generalization. At the anomaly criterion level, we derive anomaly confidence score from cluster membership probability and combine it with reconstruction error to form a dual detection criterion.

\begin{figure}[h]
    \centerline{\includegraphics[width=0.75\linewidth]{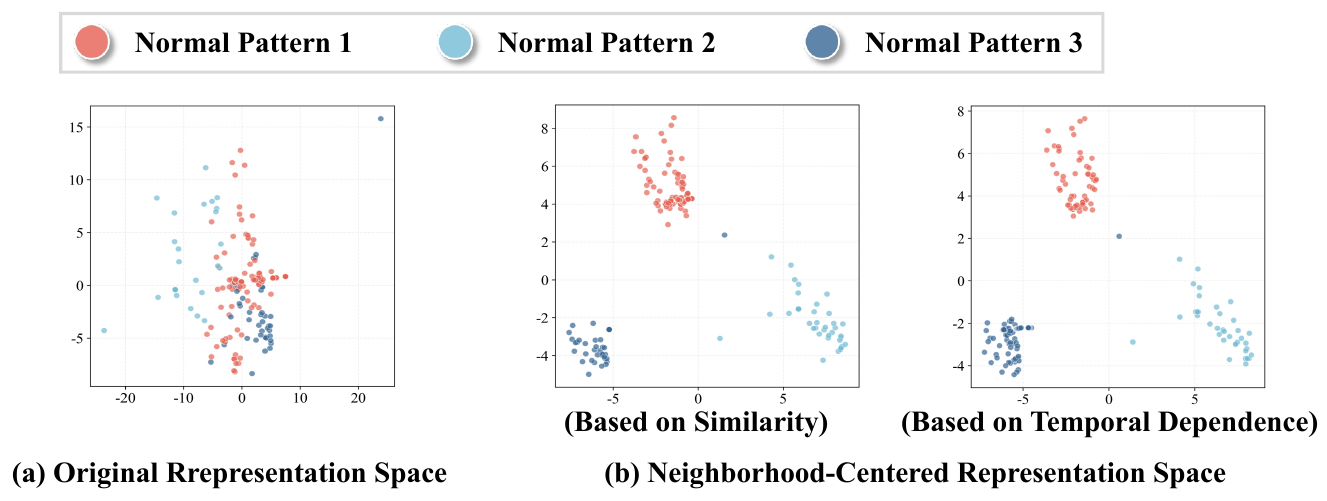}}
    \caption{Comparison of original and neighborhood-centered representation spaces. (a) Normal pattern partitioning in the original representation space, where inter-pattern boundaries are ambiguous; (b) Normal pattern partitioning in the neighborhood-centered representation space, where inter-pattern coverage is significantly reduced. All spaces are learned by SCAN on the SMAP dataset.}
    \label{fig:introduction_fig2}
\end{figure}

Second, \textbf{the effectiveness of clustering enhancement strategy depends on cluster partition quality}, although integrating clustering with reconstruction is a promising solution to Challenge 1. Shared background context across different temporal sub-patterns (e.g., basic fluctuation trends of industrial sensor data under different operating conditions) ~\cite{tan2025mask} impair accurate sample distribution partitioning, resulting in poor intra-cluster compactness and ambiguous inter-cluster boundaries, as shown in Figure ~\ref{fig:introduction_fig2}a. Low-quality clustering fails to provide reliable support for anomaly detection.
\textbf{To address this challenge}, we design a clustering module enhanced by neighborhood-centered representations. Neighborhood-centered representations indicate the degree of deviation between temporal patch representations and their neighborhood-weighted aggregated representations,
which project patch-level information into a more discriminative space, as shown in Figure~\ref{fig:introduction_fig2}b. Based on this, the differences among temporal sub-patterns can be amplified by multi-view clustering. We first extract two types of neighborhood-centered representations (based on similarity and temporal dependence respectively) from original representations as auxiliary clustering views, and perform normal pattern clustering on each view to obtain cluster centers. Then, we dynamically filter high-quality clustering pseudo-labels from neighborhood-centered representation views and use them as supervisory signals to enhance the clustering of original representations.

In summary, our contributions in this paper are as follows: 
\begin{itemize}
\item We propose a novel anomaly detection paradigm that integrates multi-scale clustering into traditional reconstruction-based methods at both the representation and anomaly criterion levels, effectively alleviating the inherent over-generalization and under-generalization issues of reconstruction-based methods.
\item We design a neighborhood-centered clustering module that extracts neighborhood-centered representations to provide trusted supervision for the clustering of original representations, thus furnishing reliable support for anomaly detection.
\item Our method outperforms state-of-the-art models and achieves superior performance on diverse datasets.
\end{itemize}

%% file: Sections/Related_Work.tex
\section{Related Work}
\label{Related_Work}

\textbf{Time Series Anomaly Detection.}
Time series anomaly detection methods are classified into traditional and deep learning methods~\cite{zhao2022comparative}. Traditional methods, including LOF~\cite{breunig2000lof}, COF~\cite{tang2002enhancing}, OCSVM~\cite{scholkopf1999support}, and SVDD~\cite{tax2004support}, detect anomalies through boundary definition or density estimation, yet fail to effectively model complex temporal dependencies. Deep learning methods adopt contrastive~\cite{xu2022anomaly,yang2023dcdetector}, forecasting~\cite{hundman2018detecting,munir2018deepant,deng2021graph,zhou2025kan}, or reconstruction paradigms. Recently, reconstruction-based methods~\cite{zong2018deep,su2019robust,zhou2019beatgan,audibert2020usad,li2021multivariate,tuli2022tranad}, such as TimesNet~\cite{wu2023timesnet}, ModernTCN~\cite{luo2024moderntcn}, and CrossAD~\cite{li2025crossad}, have shown great promise. Despite these advances, conventional reconstruction-based methods still suffer from a critical trade-off between over-generalization and under-generalization.

\textbf{Memory-based Methods. }
Memory architectures augment neural networks with external storage to retain and retrieve long-term information, with applications spanning natural language processing ~\cite{lample2019large,lewis2020retrieval,wang2023augmenting}, computer vision ~\cite{lei2020mart,oh2019video}, and other domains~\cite{le2021model,snell2017prototypical,tanwisuth2021prototype,guo2021learning}. Recently, memory-based methods have also been introduced in anomaly detection~\cite{zhou2023dual,lai2021anomaly,de2022hybrid,liu2021semi}, especially in the field of computer vision, such as MemAE~\cite{gong2019memorizing} and MNAD~\cite{park2020learning}. For time series, MEMTO~\cite{song2023memto} demonstrates the effectiveness of leveraging external memory to capture diverse normal patterns across heterogeneous domains. Nevertheless, most methods lack a reliable mechanism to amplify the differences between different temporal patterns.

%% file: Sections/Methodology.tex
\section{Methodology}
\label{Methodology}

\textbf{Problem Formulation. }
Given a multivariate time series $\mathbf{X} \in \mathbb{R}^{T \times C}$ with length $T$ and $C$ channels, the task of time series anomaly detection is to learn representations from normal training series and output labels $\mathbf{\hat{Y}}_{\text{test}} = (\hat{y}_{1}, \hat{y}_{2}, \dots, \hat{y}_{T'})$ for unseen test series $\mathbf{X}_{\text{test}}$ of length $T'$. Each label $\hat{y}_{t} \in \{0, 1\}$ indicates whether the $t$-th observation $x_{t}$ is normal or anomalous. Generally, 0 indicates normal, while 1 indicates anomalous.

\begin{figure}[h] 
    \centerline{\includegraphics[width=1\linewidth]{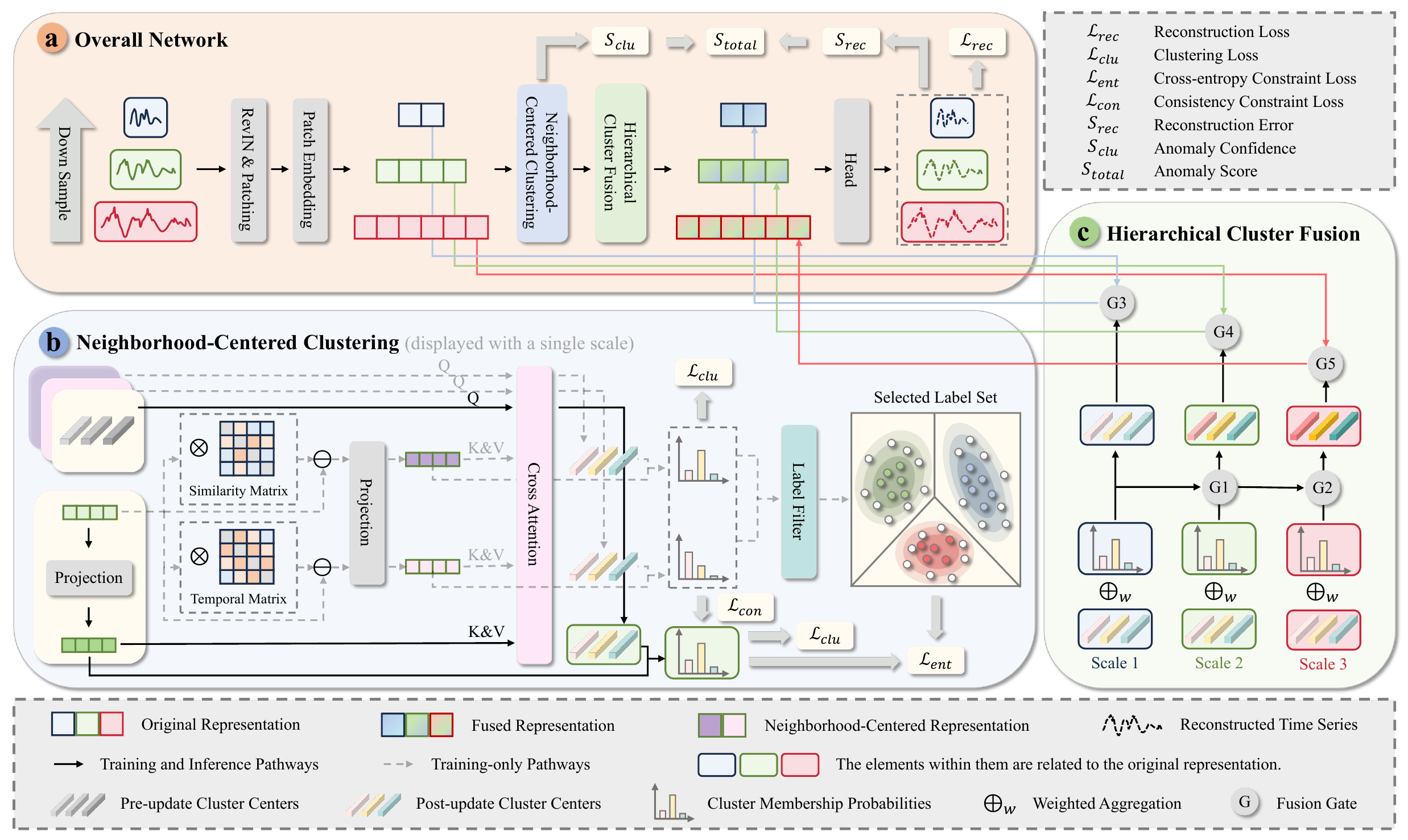}}
    \caption{The architecture of SCAN, taking scale number $j=3$ as an example.}
    \vspace{-10pt}
    \label{fig:methodology_fig1}
\end{figure}

\subsection{Overall Architecture}
The overall architecture of SCAN is shown in Figure~\ref{fig:methodology_fig1}. We first adopt multi-scale modeling to capture diverse normal patterns, including scale-specific ones: taking the raw time series $\mathbf{X}$ as $\mathbf{X}_{j}$, we apply $j$ non-overlapping average pooling operations with varying temporal kernel sizes to generate series with $j$ different scales $\{\mathbf{X}_{0}, \mathbf{X}_{1}, \dots, \mathbf{X}_{j}\}$, where the length $T_{i}$ of $\mathbf{X}_{i} \in \mathbb{R}^{T_{i} \times C}$ satisfies $T_{i} < T_{i+1}$. Each $\mathbf{X}_{i}$ undergoes reversible instance normalization, is split into $N_{i}$ patches~\cite{nie2023time}, and is projected into a $d$-dimensional space to obtain the embedding $\mathbf{E}_{i} \in \mathbb{R}^{N_{i} \times d \times C}$. These embeddings are fed into the \textit{Neighborhood-Centered Clustering Module} to extract neighborhood-centered representations. Cross-attention is then applied to cluster both neighborhood-centered and original representations, yielding normal pattern cluster centers and corresponding cluster membership probabilities. The neighborhood-centered branches generate high-quality clustering pseudo-labels through label filtering, which supervise the clustering of the original representation branch. Subsequently, in the \textit{Hierarchical Cluster Fusion Module}, cluster center representations at each scale are aggregated via a membership probability-based gating mechanism to produce scale-specific cluster-weighted representations. These are fused progressively across scales, and further combined with original representations through intra-scale fusion to form the final fused representation $\mathbf{Z}_{i} \in \mathbb{R}^{N_{i} \times d \times C}$. Finally, $\mathbf{Z}_{i}$ is mapped back to the input space via a linear projection head to reconstruct the input series. For simplicity, we will use a single-scale example to introduce the method in the following sections.

\subsection{Neighborhood-Centered Clustering}
To learn discriminative normal pattern cluster centers, we extract neighborhood-centered representations from the original ones. Their clustering processes act as auxiliary branches, offering trusted supervision for the clustering of the original representations. 

\textbf{Extraction of Neighborhood-Centered Representation. }
A neighborhood-centered representation is obtained by a weighted centering operation under the neighborhood context, which captures the deviation of each temporal patch from its neighborhood. Its extraction comprises three steps: adjacency matrix construction, neighborhood representation aggregation, and neighborhood-centered representation computation.

First, we simultaneously construct the similarity adjacency matrix $\mathbf{A}_{i}^{\mathrm{sim}} \in \mathbb{R}^{N_{i} \times N_{i}}$ and the temporal-dependent adjacency matrix $\mathbf{A}_{i}^{\mathrm{tim}} \in \mathbb{R}^{N_{i} \times N_{i}}$ to align neighborhood modeling with the data distribution.
$\mathbf{A}_{i}^{\mathrm{sim}}$ is computed by weighting flattened embeddings $\mathbf{\tilde{E}}_{i} \in \mathbb{R}^{N_{i} \times (d \times C)}$ with learnable weights $\mathbf{w} \in \mathbb{R}^{1 \times (d \times C)}$ to emphasize key features, followed by cosine similarity calculation.
$\mathbf{A}_{i}^{\mathrm{tim}}$ is initialized via patch-level temporal distances and adaptively optimized during training.
This process is formalized as:
\begin{align}
\mathbf{\tilde{E}}_i^w&=\mathcal{N}(\mathbf{\tilde{E}}_i\odot\mathbf{w}), \\
\mathbf{A}_i^{\mathrm{sim}}&=\mathrm{Softmax}(\mathbf{\tilde{E}}_i^w\cdot(\mathbf{\tilde{E}}_i^w)^\top), \\
\mathbf{A}_i^{\mathrm{tim'}}[n,m]&=\frac{1}{|n-m|+1}, \\
\mathbf{A}_i^{\mathrm{tim}}&=\mathrm{Softmax}(\mathbf{A}_i^{\mathrm{tim'}}),
\end{align}
where $\mathbf{\tilde{E}}_i^w$ denotes the weighted and normalized embedding, $\mathcal{N}$ is normalization, $\odot$ denotes element-wise multiplication, and $|n-m|$ ($0\leq|n-m|\leq N_{i}-1$) is the temporal distance between the $n$-th and $m$-th patches. 

Based on the two adjacency matrices above, we perform weighted aggregation on the neighbors of each patch to obtain two types of aggregated neighborhood representations, fusing neighborhood information. We then apply centering operations to the embedding $\mathbf{E}_{i}$ using these aggregated representations, yielding the neighborhood-centered representations $\mathbf{E}_{i}^{\mathrm{sim}} \in \mathbb{R}^{N_{i} \times d \times C}$ and $\mathbf{E}_{i}^{\mathrm{tim}} \in \mathbb{R}^{N_{i} \times d \times C}$:
\begin{align}
\mathbf{E}_i^{\mathrm{sim}}&=\mathbf{E}_i-\mathrm{Reshape}(\mathbf{A}_i^{\mathrm{sim}}\cdot\mathbf{\tilde{E}}_i), \\
\mathbf{E}_i^{\mathrm{tim}}&=\mathbf{E}_i-\mathrm{Reshape}(\mathbf{A}_i^{\mathrm{tim}}\cdot\mathbf{\tilde{E}}_i).
\end{align}

Appendix \ref{sec:Enhancement of Clustering Separability} expounds the superiority of neighborhood-centered representation over original representation.

\textbf{Pattern Clustering. }
We learn normal pattern cluster centers by clustering the normal patterns in the time series. First, the original representation $\mathbf{E}_i$ and neighborhood-centered representations $\mathbf{E}_i^{\mathrm{sim}}$, $\mathbf{E}_i^{\mathrm{tim}}$ are projected into a $d_r$-dimensional clustering space via a linear layer, producing $\mathbf{H}_i \in \mathbb{R}^{N_i \times d_r}$, $\mathbf{H}_{i}^{\mathrm{sim}} \in \mathbb{R}^{N_i \times d_r}$, and $\mathbf{H}_{i}^{\mathrm{tim}} \in \mathbb{R}^{N_i \times d_r}$, respectively. The clustering of each representation is independent. For each clustering branch, we initialize a $K$-cluster center set. Taking the original branch as an example, its cluster centers are denoted as $\mathbf{P}_{i} \in \mathbb{R}^{K \times d_r}$. For a patch representation $h_{in}$ drawn from $\mathbf{H}_i$, its cluster membership probability $u_{in,ik}$ for the $k$-th cluster $p_{ik}$ is defined as the normalized cosine similarity between $h_{in}$ and $p_{ik}$, formulated as:
\begin{align}
u_{in,ik}=\mathcal{N}\!\left(\frac{p_{ik}^{\top}h_{in}}{\|p_{ik}\|\|h_{in}\|}\right)\in[0,1],
\end{align}
where $\sum_{k=1}^K u_{in,ik}=1$.

Then, we extract normal pattern information via the masked cross-attention mechanism~\cite{chen2024similarity}:
\begin{align}
\widehat{\mathbf{P}}_{i}=\mathcal{N}(\exp(\frac{(W_Q\mathbf{P}_{i})(W_K\mathbf{H}_{i})^\top}{\sqrt{d}})\odot\mathbf{M}_{i}^\top)W_V\mathbf{H}_{i},
\end{align}
where $\widehat{\mathbf{P}}_{i}$ denotes the updated cluster center representation, and $\mathbf{M}_{i} \in \mathbb{R}^{N_i \times K}$ is the binary mask matrix, i.e., the binary cluster membership matrix obtained via Gumbel-Softmax Bernoulli sampling, with $\mathbf{M}_{in,ik} \approx \text{Bernoulli}(u_{in,ik})$. A higher $u_{in,ik}$ leads $\mathbf{M}_{in,ik}$ closer to 1. This mask guides cross-attention to focus on intra-cluster normal patterns, and thus guarantees the separability of normal pattern clusters.


Moreover, each clustering branch is optimized via a clustering loss $\mathcal{L}_{\mathrm{clu}}$, which aims to maximize intra-cluster similarity and minimize inter-cluster similarity:
\begin{align}
\mathcal{L}_{\mathrm{clu},i}^{\mathrm{ori}}
&= -\operatorname{tr}\!\left(\mathbf{M}_i^\top \mathbf{S}_i \mathbf{M}_i\right)
+ \operatorname{tr}\!\left(\left(\mathbf{I}-\mathbf{M}_i\mathbf{M}_i^\top\right)\mathbf{S}_i\right),\\
\mathcal{L}_{\mathrm{clu}}
&=\sum_{i=1}^{m}\left(\mathcal{L}_{\mathrm{clu},i}^{\mathrm{ori}}
+\mathcal{L}_{\mathrm{clu},i}^{\mathrm{sim}}
+\mathcal{L}_{\mathrm{clu},i}^{\mathrm{tim}}\right),
\end{align}
where $\mathbf{S}_i\in\mathbb{R}^{N_i\times N_i}$ is the same-cluster probability matrix constructed using K-means in the clustering representation space. Specifically, we derive the soft assignment $\mathbf{Y}_i^{\mathrm{soft}}[n,k]$ that patch $n$ belongs to cluster $k$ via distance-based softmax, and $S_i[n,m] = \sum_{k=1}^{K} \mathbf{Y}_i^{\mathrm{soft}}[n,k] \mathbf{Y}_i^{\mathrm{soft}}[m,k]$ measures the probability that $n$-th patch and $m$-th patch belong to the same cluster.

\textbf{Trusted Supervision. }
\label{sec:Trusted Supervision}
Clustering of the original representation is prone to interference from inter-cluster shared background context, leading to a performance bottleneck. To address this, we enhance it using more discriminative clustering views based on neighborhood-centered representations. We further use a filtering mechanism to select high-quality pseudo-labels of the neighborhood-centered representation branches as supervisory signals for the clustering of original representation~\cite{li2022twin}.

In pattern clustering, each branch outputs a clustering result, where each patch corresponds to three cluster membership probability vectors.
For the $n$-th patch, the membership vectors from the neighborhood-centered representation branches are denoted as
$\mathbf{u}_{in}^{\mathrm{sim}}=\{u_{in,i1}^{\mathrm{sim}},\dots,u_{in,iK}^{\mathrm{sim}}\}$ and $\mathbf{u}_{in}^{\mathrm{tim}}=\{u_{in,i1}^{\mathrm{tim}},\dots,u_{in,iK}^{\mathrm{tim}}\}$, satisfying $\sum_{k=1}^{K}u_{in,ik}^{(\cdot)}=1$ with $(\cdot)\in\{\mathrm{sim},\mathrm{tim}\}$.



To measure the reliability of cluster assignments, we utilize the properties of Shannon entropy for the membership vector $\mathbf{u} \in \mathbb{R}^K$. It is well-established that the entropy $H(\mathbf{u}) = -\sum_{k=1}^{K} u_k \log u_k$ reaches its maximum, $H_{\max} = \log K$, when $\mathbf{u}$ follows a uniform distribution, representing the highest uncertainty. 

Based on this, we define the information redundancy as $R(\mathbf{u}) = 1 - H(\mathbf{u})/H_{\max}$. To selectively utilize highly certain pseudo-labels, we adopt an entropy-based quality score:
\begin{equation}
\mathrm{Quality}(\mathbf{u}) = 1 - \frac{H(\mathbf{u})}{2H_{\max}},
\end{equation}
where a higher $\mathrm{Quality}(\mathbf{u})$ indicates a more peaked distribution and thus a more reliable cluster assignment.

For each neighborhood-centered representation branch, patches are first ranked by their maximum membership probability $\max_k u_{in,ik}^{(\cdot)}$.
We then dynamically determine the number of selected pseudo-labeled patches by summing the quality scores, and choose the top $B_{i}^{\mathrm{sel},(\cdot)}=\left\lfloor \sum_{n=1}^{N_i} \mathrm{Quality}(\mathbf{u}_{in}^{(\cdot)}) \right\rfloor$ patches to form the trusted index set $\Omega_i^{(\cdot)}$ for the branch.
The pseudo-label for each selected patch is defined as $\tilde{c}_{in}^{(\cdot)}=\arg\max_{k} u_{in,ik}^{(\cdot)}, \quad n\in\Omega_i^{(\cdot)}$.
The trusted pseudo-labels are used to supervise the cluster membership probabilities $\mathbf{u}_{in}=\{u_{in,i1},\dots,u_{in,iK}\}$ of the original representation branch.
Specifically, we minimize the cross-entropy constraint loss $\mathcal{L}_{\mathrm{ent}}$ between the original cluster assignments and the filtered pseudo-labels:
\begin{align}
\mathcal{L}_{\mathrm{ent}} = \sum_{i=1}^{m}\left(-\frac{1}{|\Omega_i|}\sum_{n\in\Omega_i}\phi(\tilde{c}_{in})\log{u_{in}}\right),
\end{align}
where $\phi$ denotes the one-hot encoding function.

Moreover, we use the consistency constraint loss $\mathcal{L}_{\mathrm{con}}$ to enforce consistency between the neighborhood-centered representations and the original representations in the semantic space:
\begin{align}
\mathcal{L}_\mathrm{con}=\sum_{i=1}^{m} D_{\mathrm{KL}}(\mathbf{U}_i^{(\cdot)}\parallel \mathbf{U}_i), \quad (\cdot) \in \{\mathrm{sim},\mathrm{tim}\},
\end{align}
where $\mathbf{U}_i \in \mathbb{R}^{N_i \times K}$ denotes the membership probability matrix.

\subsection{Hierarchical Cluster Fusion}
To ensure the model target representative normal patterns for reconstruction, maintaining resistance to over-generalization while preserving key details of the original representation, we integrate cluster center representations into the representations used for reconstruction.

\textbf{Inter-scale Fusion. }
After obtaining the scale-specific cluster centers, we derive a cluster-weighted representation for each patch by weighting the cluster center representations with the corresponding cluster membership probabilities.
For the $i$-th scale, given the updated cluster center representations $\widehat{\mathbf{P}}_{i} \in \mathbb{R}^{K \times d_r}$ and the membership probability matrix $\mathbf{U}_i \in \mathbb{R}^{N_i \times K}$, the cluster-weighted representation $\mathbf{G}_i \in \mathbb{R}^{N_i \times d_r}$ is computed as $\mathbf{G}_i=\mathbf{U}_i \cdot \widehat{\mathbf{P}}_{i}$.

To propagate normal patterns from coarser to finer scales, we conduct progressive inter-scale fusion via gating mechanisms. For the coarsest scale, we directly set $\mathbf{F}_0=\mathbf{G}_0$. For a finer scale $i>0$, we first align the previous fused cluster-weighted representation $\mathbf{F}_{i-1}$ to the resolution of the current scale via linear interpolation to obtain $\widetilde{\mathbf{F}}_{i-1}$, then fuse it with $\mathbf{G}_i$ using a learnable gate:
\begin{align}
\boldsymbol{\alpha}_i &= \sigma\!(\mathrm{Gate}_i^{\mathrm{inter}}[\mathbf{G}_i;\widetilde{\mathbf{F}}_{i-1}]), \\
\mathbf{F}_i &= \boldsymbol{\alpha}_i \odot \mathbf{G}_i + (1-\boldsymbol{\boldsymbol{\alpha}}_i) \odot \widetilde{\mathbf{F}}_{i-1},
\end{align}
where $[\cdot;\cdot]$ denotes concatenation and $\sigma(\cdot)$ denotes the sigmoid function. This design enables the model to adaptively balance scale-specific details and cross-scale global context.

\textbf{Intra-scale Fusion. }
The fused cluster-weighted representation $\mathbf{F}_i\in\mathbb{R}^{N_i\times d_r}$ is further integrated with the original patch embedding $\mathbf{E}_i$ to construct the final representation for reconstruction.
As $\mathbf{F}_i$ lies in the cluster representation space while $\mathbf{E}_i$ resides in the reconstruction space, we first project $\mathbf{F}_i$ to $\mathbf{E}_i^{\mathrm{clu}} \in \mathbb{R}^{N_i \times (d \times C)}$ via a linear layer. A learnable gate is then used to fuse $\mathbf{E}_i$ and $\mathbf{E}_i^{\mathrm{clu}}$:
\begin{align}
\boldsymbol{\beta}_i &= \sigma\!\left(\mathrm{Gate}_i^{\mathrm{intra}}[\mathbf{E}_i;\mathbf{E}_i^{\mathrm{clu}}]\right), \\
\mathbf{Z}_i &= \boldsymbol{\beta}_i \odot \mathbf{E}_i + (1-\boldsymbol{\beta}_i)\odot \mathbf{E}_i^{\mathrm{clu}},
\end{align}
where $\mathbf{Z}_i$ denotes the final fused representation for reconstruction. By preserving the original embedding and injecting a clustering-guided normal pattern constraint, this intra-scale fusion alleviates both over-generalization to anomalies and under-generalization to normal patterns. 

Finally, we minimize the Mean Squared Error (MSE) between the input series and its reconstructed result:
\begin{align}
\mathcal{L}_{\mathrm{rec}}
= \sum_{i=1}^{m}||\mathbf{X}_i-\mathbf{\hat{X}}_i||_2^2.
\end{align}

\subsection{Loss Function}
\label{sec:Loss Function}
The objective function of SCAN contains four components: reconstruction loss $\mathcal{L}_{\mathrm{rec}}$, clustering loss $\mathcal{L}_{\mathrm{clu}}$, cross-entropy constraint loss $\mathcal{L}_{\mathrm{ent}}$, and consistency constraint loss $\mathcal{L}_{\mathrm{con}}$.
Thus, the total loss $\mathcal{L}_{\mathrm{total}}$ is defined as follows:
\begin{align}
\mathcal{L}_\mathrm{total} = \mathcal{L}_\mathrm{rec} + \lambda_{1} \mathcal{L}_\mathrm{clu} + \lambda_{2} \mathcal{L}_\mathrm{ent} + \lambda_{3} \mathcal{L}_\mathrm{con},
\end{align}
where $\lambda_{1}$, $\lambda_{2}$, and $\lambda_{3}$ are the balance coefficients.

\subsection{Anomaly Criterion}
The anomaly score $\mathcal{S}_\mathrm{total}$ of SCAN consists of two components: reconstruction error $\mathcal{S}_\mathrm{rec}$ and anomaly confidence $\mathcal{S}_\mathrm{clu}$. We first interpolate the scale-wise anomaly scores with varying lengths to align them to a uniform length $T_m$:
\begin{align}
\mathcal{S}_\mathrm{rec} &= \prod_{i=1}^{m}\mathrm{interpolate}\!((\mathbf{X}_i-\hat{\mathbf{X}}_{i})^2), \\
\mathcal{S}_\mathrm{clu} &= \prod_{i=1}^{m}\mathrm{interpolate}\!(1-\max_{k\in\{1,\dots,K\}}\mathbf{U}_i(:,k)), \\
\mathcal{S}_\mathrm{total} &= (\mathcal{S}_\mathrm{rec})^{1-\gamma} \cdot (\mathcal{S}_\mathrm{clu})^{\gamma},
\end{align}
where $\gamma$ is the balancing coefficient and $\hat{\mathbf{X}}_{i}$ is the reconstructed results. Following prior works, we run the SPOT~\cite{siffer2017anomaly} to automatically compute the threshold $\delta$ after obtaining anomaly score. A point is marked as an anomaly if its anomaly score exceeds $\delta$.

%% file: Sections/Experiments.tex
\section{Experiments}
\label{Experiments}

\subsection{Experimental Setup}

\textbf{Datasets. }
We evaluate on various widely adopted benchmark datasets, including SMD~\cite{su2019robust}, MSL~\cite{hundman2018detecting}, SMAP~\cite{hundman2018detecting}, PSM~\cite{abdulaal2021practical}, SWAT~\cite{mathur2016swat}, and NeurIPS-TS~\cite{lai2021revisiting} (which encompasses GECCO and SWAN). We also conduct experiments on the UCR dataset~\cite{wu2021current}, with corresponding results presented in Appendix~\ref{sec:UCR Benchmark}. 

\textbf{Baselines. }
We extensively compare our model with 22 baselines, including the latest state-of-the-art (SOTA) anomaly detection models. These baselines comprise: linear transformation-based methods: OCSVM~\cite{scholkopf1999support}, PCA~\cite{shyu2003novel}; density estimation-based methods: LOF~\cite{breunig2000lof}, HBOS~\cite{goldstein2012histogram}; outlier-based methods: IForest~\cite{liu2008isolation}, LODA~\cite{pevny2016loda}; neural network-based methods: Autoencoders (AE)~\cite{sakurada2014anomaly}, DAGMM~\cite{zong2018deep}, LSTM~\cite{hundman2018detecting}, OmniAnomaly (Omni)~\cite{su2019robust}, CAE Ensemble (CAE)~\cite{campos2021unsupervised}, Anomaly Transformer (AT)~\cite{xu2022anomaly}, TimesNet~\cite{wu2023timesnet}, DC Detector (DC)~\cite{yang2023dcdetector}, GPT4TS~\cite{zhou2023one}, ModernTCN~\cite{luo2024moderntcn}, TimeMixer~\cite{wangtimemixer}, MtsCID~\cite{xie2025multivariate}, MEMTO~\cite{song2023memto}, DADA~\cite{shentu2025towards}, CrossAD~\cite{li2025crossad}, KAN-AD~\cite{zhou2025kan}.

\textbf{Metrics. }
Many methods refine detection results via point adjustment~\cite{yang2023dcdetector,xu2022anomaly}. However, this strategy incorrectly assumes that a single accurate point detection validates an entire anomalous segment. To address this, some works adopt the Affiliation-F1~\cite{huet2022local}, yet it is highly threshold-sensitive. Recent studies confirm VUS-PR~\cite{paparrizos2022volume,boniol2025vus} as the most robust, accurate and fair metric. We thus mainly use VUS-ROC and VUS-PR, alongside mainstream metrics for comprehensive comparison, which are provided in Appendix~\ref{sec:Multi-metrics Results}. Further implementation details are presented in Appendix~\ref{sec:Implementation Details}.

\subsection{Main Results}
We evaluate SCAN on seven real-world datasets against 22 baselines, as shown in Table~\ref{tab:table1}. SCAN achieves state-of-the-art performance under both VUS-ROC (V-R) and VUS-PR (V-P), exhibiting superior detection accuracy and stability over various pre-selected thresholds. This validates the effectiveness of our neighborhood-centered clustering and hierarchical cluster fusion, which refine reconstruction representations and produce clustering-based anomaly confidence scores, offering a new paradigm for time series anomaly detection.

\input{Tables/table1}


\subsection{Model Analysis}
\textbf{Ablation Studies. }
To validate the effectiveness of SCAN, we conduct ablation studies on its key modules, as shown in Table~\ref{tab:table3}. Multi-scale modeling consistently improves model performance. Based on this, pattern clustering enables reconstruction with cluster center representations, further boosting metrics. Applying trusted supervision to pattern clustering increases VUS-ROC and VUS-PR by 1.49\% and 9.42\%, respectively, owing to effective anomaly confidence scores. Integrating cluster center representations into original features improves the two metrics by 2.01\% and 8.63\%, respectively, demonstrating the necessity of representation fusion. Moreover, removing the multi-scale cluster fusion module degrades performance, verifying such fusion enriches normal pattern modeling.
\input{Tables/table3}

\begin{wrapfigure}{r}{0.6\linewidth}
  \centering

  \caption{Static and runtime performance metrics of SCAN and other time series anomaly detection  methods on the GECCO dataset, evaluated with train epochs of 5 and a batch size of 128.}
  \vspace{3mm}
  \rowcolors{2}{white}{gray!10}
  \resizebox{1.0\linewidth}{!}{
    \setlength{\tabcolsep}{4pt}
    \begin{tabular}{c|c|c|c}
    \toprule
    Method & Training Time (s) & Inference Time (s) & Total Params (M) \\
    \midrule
    ModernTCN & 22.75 & 0.57  & \textbf{0.05} \\
    TimesNet & 342.79 & 1.68  & 4.68 \\
    TimeMixer & 57.2  & 0.87  & 0.1 \\
    AnomalyTransformer & 270.49 & 15.05 & 4.74 \\
    CrossAD & 270.75 & 0.80 & 0.93 \\
    KAN-AD & 68.92 & \textbf{0.12} & 0.41 \\
    \midrule
    \rowcolor{red!10}
    SCAN  & \textbf{15.11} & 0.15  & 1.76 \\
    \bottomrule
    \end{tabular}}%
  \label{tab:table4}%
  \vspace{20pt}
  \includegraphics[width=\linewidth]{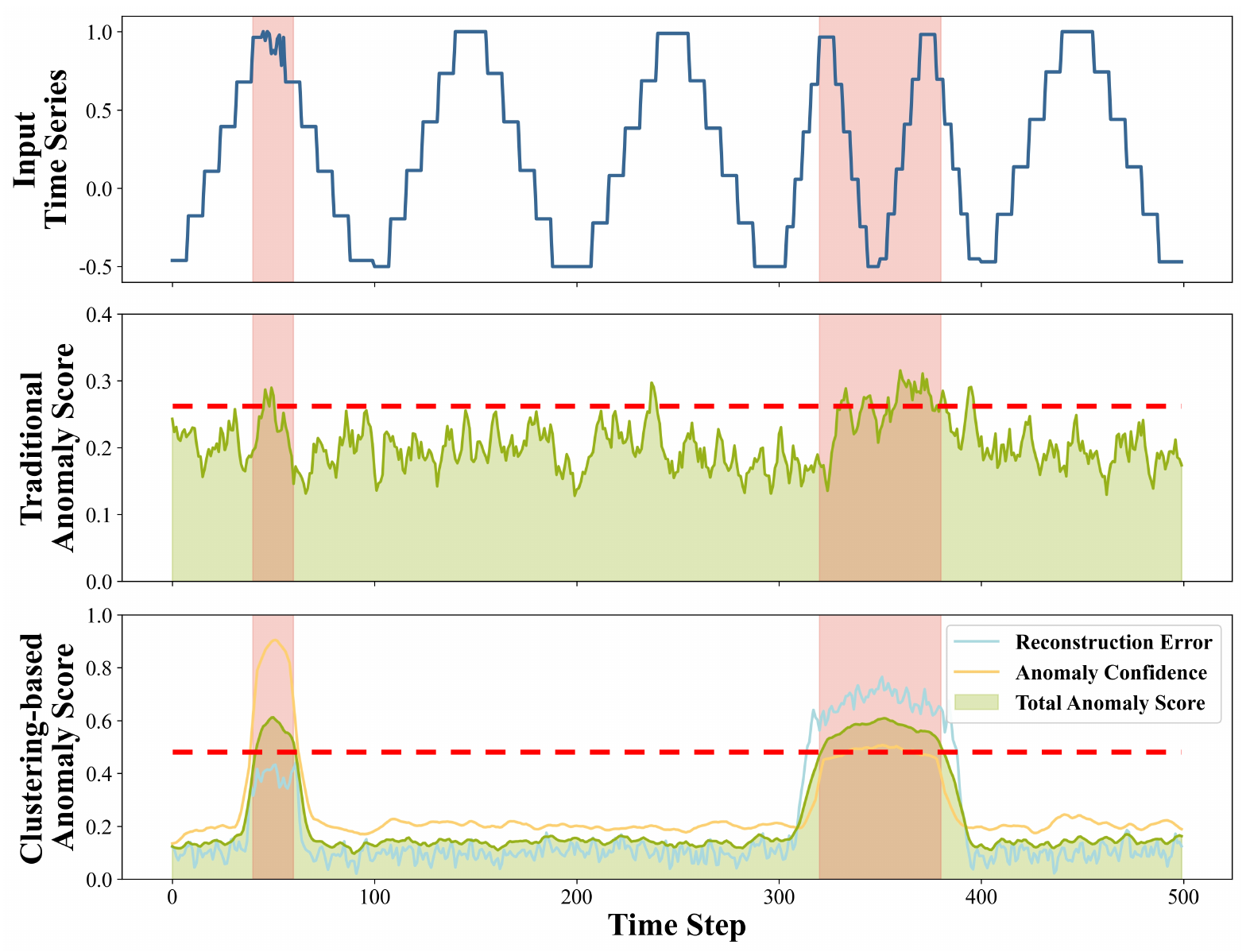}
  \vspace{-10pt}
  \caption{Visualization of traditional and clustering-based anomaly scores. From top to bottom: time series, traditional anomaly scores, clustering-based anomaly scores. Red regions mark anomalies.}
  \vspace{-30pt}
  \label{fig:experiments_fig1}
\end{wrapfigure}

\textbf{Model Efficiency. }
We compare the efficiency of SCAN with representative time series anomaly detection methods, including MLP-based (TimeMixer), CNN-based (ModernTCN, TimesNet, KAN-AD) and Transformer-based (AnomalyTransformer, CrossAD) methods, using both static and runtime metrics, as shown in Table~\ref{tab:table4}. SCAN achieves superior detection performance alongside speed advantage. 
In addition, the time complexity of SCAN is 
$\mathcal{O}\left( \sum_{i=1}^{m} \Big( N_i^2 (d_{\text{emb}} + K) + \right.$ 
$\left. N_i (d_{\text{emb}}d_r + Kd_r + Ld_{emb}) \Big) \right)$ 
with controllable overhead, where $d_{\text{emb}}$ denotes the embedding dimension ($d_{\text{emb}}=d \times C$), and $L$ represents the patch size. The detailed complexity analysis is presented in Appendix~\ref{sec:Complexity Analysis}.

\textbf{Visualization. }
To validate the efficacy of clustering for anomaly detection, we perform a comparative visualization of clustering-based and traditional anomaly scores, as shown in Figure~\ref{fig:experiments_fig1}. The time series contains two complex anomalies: shapelet and seasonal. Traditional criterion exhibits high false positive and negative rates, while clustering-based criterion, which integrates reconstruction error and anomaly confidence scores, enables stable and accurate detection.



%% file: Tables/table1.tex
\begin{table}[htbp]
  \centering
  \renewcommand{\arraystretch}{1.1}
  \caption{VUS-metrics results in the seven real-world datasets. Higher VUS-ROC (V-R) or VUS-PR (V-P) values indicate better performance. \textcolor[rgb]{ 1,  0,  0}{\textbf{Red}}: the best, \textcolor[rgb]{ .161,  .447,  .957}{\underline{Blue}}: the 2nd best.}
  \rowcolors{2}{gray!10}{white}
  \resizebox{1.0\linewidth}{!}{
    \begin{tabular}{c|cc|cc|cc|cc|cc|cc|cc}
    \toprule
    Dataset & \multicolumn{2}{c|}{SMD} & \multicolumn{2}{c|}{MSL} & \multicolumn{2}{c|}{SMAP} & \multicolumn{2}{c|}{PSM} & \multicolumn{2}{c|}{SWaT} & \multicolumn{2}{c|}{GECCO} & \multicolumn{2}{c}{SWAN} \\
    \midrule
    Metric & V-R   & V-P   & V-R   & V-P   & V-R   & V-P   & V-R   & V-P   & V-R   & V-P   & V-R   & V-P   & V-R   & V-P \\
    \midrule
    OCSVM & 0.6451  & 0.1131  & 0.5798  & 0.1753  & 0.4185  & 0.1133  & 0.5993  & 0.4252  & 0.5903  & 0.4396  & 0.7533  & 0.1207  & 0.9088  & 0.9004  \\
    PCA   & 0.7174  & 0.1529  & 0.6108  & 0.1889  & 0.4090  & 0.1144  & 0.6331  & 0.4706  & 0.6149  & 0.4459  & 0.5366  & 0.0443  & 0.9290  & 0.9123  \\
    IForest & 0.7224  & 0.1304  & 0.5638  & 0.1631  & 0.4960  & 0.1315  & 0.6009  & 0.3964  & 0.3677  & 0.1011  & 0.7083  & 0.0943  & 0.8835  & 0.8793  \\
    LODA  & 0.6745  & 0.1213  & 0.5375  & 0.1689  & 0.3973  & 0.1017  & 0.6089  & 0.4423  & 0.6358  & 0.3531  & 0.5749  & 0.0339  & 0.9170  & 0.9107  \\
    HBOS  & 0.6670  & 0.1102  & 0.6265  & 0.1790  & 0.5620  & 0.1388  & 0.7056 & 0.5061 & 0.7084  & 0.4602  & 0.5440  & 0.0453  & 0.9056  & 0.8894  \\
    LOF   & 0.6893  & 0.1076  & 0.6081  & 0.1715  & 0.5673 & 0.1409 & 0.6628  & 0.4615  & 0.6667  & 0.4187  & 0.7817  & 0.0919  & 0.9095  & 0.9007  \\
    AE    & 0.7560  & 0.1542  & 0.6047  & 0.1890  & 0.4687  & 0.1366  & 0.6339  & 0.4490  & 0.5903  & 0.4144  & 0.6124  & 0.0448  & 0.6982  & 0.0201  \\
    DAGMM & 0.6988  & 0.1496  & 0.6069  & 0.1803  & 0.5599  & 0.1349  & 0.5598  & 0.4522  & 0.5746  & 0.4731  & 0.5099  & 0.0396  & 0.8951  & 0.8697  \\
    LSTM  & 0.7001  & 0.1395  & 0.6163  & 0.1681  & 0.5329  & 0.1399  & 0.5571  & 0.4592  & 0.5482  & 0.2200  & 0.6450  & 0.0668  & 0.9082  & 0.8862  \\
    CAE   & 0.7174  & 0.1376  & 0.5382  & 0.1639  & 0.4212  & 0.1140  & 0.6113  & 0.4395  & 0.5939  & 0.4104  & 0.5524  & 0.0528  & 0.9042  & 0.9022  \\
    Omni  & 0.7080  & 0.1340  & 0.5490  & 0.1973  & 0.4743  & 0.1239  & 0.6340  & 0.4472  & 0.6187  & 0.4475  & 0.5386  & 0.0517  & 0.9041  & 0.9022  \\
    AT    & 0.5117  & 0.0796  & 0.3890  & 0.1041  & 0.4571  & 0.1239  & 0.5186  & 0.3309  & 0.5561  & 0.2679  & 0.4751  & 0.0278  & 0.8046  & 0.7943  \\
    DC    & 0.5145  & 0.0814  & 0.3900  & 0.0948  & 0.4444  & 0.1149  & 0.5235  & 0.3366  & 0.5191  & 0.1495  & 0.5454  & 0.0361  & 0.8429  & 0.8338  \\
    GPT4TS & 0.7679  & 0.1745  & 0.7697  & 0.2769  & 0.5449  & 0.1289  & 0.6466  & 0.4599  & 0.2537  & 0.0846  & 0.9776  & 0.4181  & 0.9340  & 0.8924  \\
    ModernTCN & 0.7707  & 0.1596  & 0.7747  & 0.3010 & 0.5470  & 0.1395  & 0.6480  & 0.4668  & 0.2735  & 0.0941  & 0.9694  & 0.4819 & 0.9027  & 0.8962  \\
    MtsCID & 0.5162  & 0.0815  & 0.4686  & 0.1181  & 0.4260  & 0.1177  & 0.5194  & 0.3296  & 0.5021  & 0.1283  & 0.5315  & 0.0381  & 0.8128  & 0.8375  \\
    TimeMixer & 0.7711  & 0.1391  & 0.7858  & 0.2461  & 0.5552  & 0.1371  & 0.5974  & 0.3807  & 0.2673  & 0.0918  & 0.9899 & 0.4606  & 0.9290  & 0.8721  \\
    TimesNet & 0.8420 & 0.2040 & 0.7880 & 0.2731  & 0.5495  & 0.1352  & 0.6344  & 0.4373  & 0.2974  & 0.1158  & 0.9834  & 0.4578  & 0.9515  & 0.9160 \\
    MEMTO & 0.5165  & 0.0824  & 0.5077  & 0.1521  & 0.5059  & 0.1416  & 0.5208  & 0.3307  & 0.7354  & 0.2858  & 0.6018  & 0.0449  & 0.8158  & 0.8389  \\
    DADA  & 0.8276  & 0.1624  & 0.7918  & 0.3028  & 0.4535  & 0.1223  & 0.6940  & 0.4782  & 0.7714  & 0.4422  & 0.9791  & 0.4588  & \textcolor[rgb]{ .161,  .447,  .957}{\underline{0.9521 }} & 0.9124  \\
    CrossAD & \textcolor[rgb]{ .161,  .447,  .957}{\underline{0.8580}} & \textcolor[rgb]{ .161,  .447,  .957}{\underline{0.2344}} & \textcolor[rgb]{ .161,  .447,  .957}{\underline{0.8091}} & \textcolor[rgb]{ .161,  .447,  .957}{\underline{0.3144}} & \textcolor[rgb]{ .161,  .447,  .957}{\underline{0.5779}} & \textcolor[rgb]{ .161,  .447,  .957}{\underline{0.1443}} & \textcolor[rgb]{ .161,  .447,  .957}{\underline{0.7302}} & \textcolor[rgb]{ .161,  .447,  .957}{\underline{0.5596}} & \textcolor[rgb]{ .161,  .447,  .957}{\underline{0.7865}} & 0.4767  & \textcolor[rgb]{ .161,  .447,  .957}{\underline{0.9948}} & \textcolor[rgb]{ .161,  .447,  .957}{\underline{0.6211}} & 0.9499  & \textcolor[rgb]{ .161,  .447,  .957}{\underline{0.9171}} \\
    KAN-AD & 0.7657  & 0.1593  & 0.6629  & 0.2107  & 0.4772  & 0.1276  & 0.6377  & 0.4523  & 0.7821 & \textcolor[rgb]{ 1,  0,  0}{\textbf{0.5829}} & 0.8567  & 0.1732  & 0.4433  & 0.3909  \\
    \midrule
    \rowcolor{red!10}
    SCAN  & \textcolor[rgb]{ 1,  0,  0}{\textbf{0.9081}} & \textcolor[rgb]{ 1,  0,  0}{\textbf{0.3203}} & \textcolor[rgb]{ 1,  0,  0}{\textbf{0.8232}} & \textcolor[rgb]{ 1,  0,  0}{\textbf{0.3315}} & \textcolor[rgb]{ 1,  0,  0}{\textbf{0.6114}} & \textcolor[rgb]{ 1,  0,  0}{\textbf{0.1578}} & \textcolor[rgb]{ 1,  0,  0}{\textbf{0.7457}} & \textcolor[rgb]{ 1,  0,  0}{\textbf{0.5851}} & \textcolor[rgb]{ 1,  0,  0}{\textbf{0.7866}} & \textcolor[rgb]{ .161,  .447,  .957}{\underline{0.4986}} & \textcolor[rgb]{ 1,  0,  0}{\textbf{0.9957}} & \textcolor[rgb]{ 1,  0,  0}{\textbf{0.6812}} & \textcolor[rgb]{ 1,  0,  0}{\textbf{0.9591}} & \textcolor[rgb]{ 1,  0,  0}{\textbf{0.9233}} \\
    \bottomrule
    \end{tabular}}%
  \label{tab:table1}%
  \vspace{-10pt}
\end{table}%

%% file: Tables/table3.tex
\begin{table}[htbp]
  \centering
  \renewcommand{\arraystretch}{1.1}
  \caption{Ablations on the key components of SCAN, including Multi-Scale Modeling, Pattern Clustering, Trusted Supervision, and Cluster Fusion. Higher VUS-ROC (V-R) or VUS-PR (V-P) values indicate better performance. \textbf{Bold}: the best.}
  \rowcolors{4}{white}{gray!10}
  \resizebox{1.0\linewidth}{!}{
    \begin{tabular}{cccc|cc|cc|cc|rr}
    \toprule
    \multicolumn{4}{c|}{Dataset}  & \multicolumn{2}{c|}{PSM} & \multicolumn{2}{c|}{SMAP} & \multicolumn{2}{c|}{GECCO} & \multicolumn{2}{c}{Avg.} \\
    \midrule
    Multi-Scale & Pattern & Trusted & Cluster & \multirow{2}[2]{*}{V-R} & \multirow{2}[2]{*}{V-P} & \multirow{2}[2]{*}{V-R} & \multirow{2}[2]{*}{V-P} & \multirow{2}[2]{*}{V-R} & \multirow{2}[2]{*}{V-P} & \multicolumn{1}{c}{\multirow{2}[2]{*}{V-R}} & \multicolumn{1}{c}{\multirow{2}[2]{*}{V-P}} \\
    Modeling & Clustering & Supervision & Fusion &       &       &       &       &       &       &       &  \\
    \midrule
    \xmark     & \xmark     & \xmark     & \xmark     & 0.6754  & 0.4638  & 0.5067  & 0.1312  & 0.9767  & 0.4542  & 0.7196  & 0.3497  \\
    \cmark     & \xmark     & \xmark     & \xmark     & 0.6901  & 0.4936  & 0.5348  & 0.1375  & 0.9817  & 0.5096  & 0.7356  & 0.3802  \\
    \cmark     & \cmark     & \xmark     & \xmark     & 0.7008  & 0.5118  & 0.5876  & 0.1495  & 0.9842  & 0.5368  & 0.7575  & 0.3994  \\
    \cmark     & \cmark     & \cmark     & \xmark     & 0.7185  & 0.5416  & 0.5963  & 0.1528  & 0.9916  & 0.6167  & 0.7688  & 0.4370  \\
    \xmark     & \cmark     & \cmark     & \xmark     & 0.7090  & 0.5278  & 0.5767  & 0.1477  & 0.9879  & 0.5742  & 0.7579  & 0.4165  \\
    \cmark     & \cmark     & \cmark     & \cmark     & \textbf{0.7457} & \textbf{0.5851} & \textbf{0.6114} & \textbf{0.1578} & \textbf{0.9957} & \textbf{0.6812} & \textbf{0.7843}  & \textbf{0.4747}  \\
    \bottomrule
    \end{tabular}}%
  \label{tab:table3}%
  \vspace{-10pt}
\end{table}%

%% file: Sections/Conclusion.tex
\section{Conclusion}
\label{Conclusion}


In this work, we propose SCAN, a clustering-enhanced time series anomaly detection model. By integrating clustering and reconstruction within a novel paradigm, and leveraging neighborhood-centered representations to boost clustering, we mitigate the over-generalization and under-generalization trade-off dilemma in reconstruction-based methods. However, like most clustering methods, SCAN requires a predefined number of clusters, which limits its flexibility. In future work, we plan to evaluate more real-world scenarios and explore adaptive clustering algorithms.

\clearpage

%% file: Sections/Appendix.tex
\appendix

\section{Analysis: Enhancement of Clustering Separability}
\label{sec:Enhancement of Clustering Separability}

To analyze why neighborhood-centered representation can improve clustering, we model each patch embedding as the sum of a specific pattern and a locally shared background context. Based on this model, we state a proposition about the expected change in clustering contrast when we apply neighborhood centering.

\begin{definition}[Additive Shared-Context Model]
\label{def:additive_model}
Let the patch embedding $\mathbf{e}_{in}$ (the $n$-th patch at the $i$-th scale) be expressed as
$\mathbf{e}_{in} = \mathbf{a}_{in} + \mathbf{b}_{in}$,
where $\mathbf{a}_{in}$ denotes a specific pattern component and $\mathbf{b}_{in}$ denotes a locally shared background context component.
We define the background-to-pattern intensity ratio as
$\alpha = \frac{\|\mathbf{b}_{in}\|_2^2}{\|\mathbf{a}_{in}\|_2^2}$ $(\alpha \ge 0)$.

Within a local neighborhood $\mathcal{N}_{in}$,
the background context changes slowly in expectation.
Formally, centering by neighborhood averaging mainly reduces the contribution of $\mathbf{b}_{in}$ to similarity computations.
\end{definition}

\begin{definition}[Clustering Contrast]
\label{def:clustering_contrast}
We define the Clustering Contrast $\Delta$ as the expected margin between intra-cluster similarity and inter-cluster similarity:
\begin{equation}
\begin{aligned}
\Delta ={}& \mathbb{E}[\mathrm{Sim}(\mathbf{e}_{in}, \mathbf{e}_{im}) \mid p_{in} = p_{im}]
- \mathbb{E}[\mathrm{Sim}(\mathbf{e}_{in}, \mathbf{e}_{im}) \mid p_{in} \neq p_{im}],
\end{aligned}
\end{equation}
where $p_{in}$ and $p_{im}$ denote the clusters of $e_{in}$ and $e_{im}$ relatively, and $\mathrm{Sim}(\cdot, \cdot)$ denotes cosine similarity.
A larger $\Delta$ indicates a more separable representation space.
\end{definition}

\begin{proposition}[Separability Enhancement]
\label{prop:separability_enhancement}
Under the Additive Shared-Context Model (Definition \ref{def:additive_model}), applying neighborhood centering to obtain a neighborhood-centered representation
reduces the influence of the shared background context more strongly than it reduces the specific pattern contribution.
As a result, the clustering contrast of the neighborhood-centered representation ($\Delta_{\text{NCR}}$) is improved compared to that of the original representation ($\Delta_{\text{OR}}$),
and the improvement is approximately proportional to $(1+\alpha)$.
\end{proposition}

\section{Implementation Details}
\label{sec:Implementation Details}

\subsection{Datasets}
\label{sec:Datasets}
In order to verify the effectiveness of our proposed model, we carried out extensive experiments on a range of publicly accessible datasets. These datasets were chosen due to their diverse data characteristics and anomaly types, thereby forming a comprehensive evaluation framework. The selected datasets include SMD (Server Machine Dataset)~\cite{su2019robust}, MSL (Mars Science Laboratory Dataset)~\cite{hundman2018detecting}, SMAP (Soil Moisture Active Passive Dataset)~\cite{hundman2018detecting}, PSM (Pooled Server Metrics Dataset)~\cite{abdulaal2021practical}, SWaT (Secure Water Treatment)~\cite{mathur2016swat}, the GECCO and SWAN sub-datasets from NeurIPS-TS (NeurIPS 2021 Time Series Benchmark)~\cite{lai2021revisiting}, and UCR~\cite{wu2021current}. Detailed descriptions of each dataset are provided below:
\begin{enumerate}
\item SMD collects resource utilization information from the computer clusters owned by an Internet company. 
\item MSL, collected by NASA, contains telemetry data that reflects the operational conditions of sensors and actuators on the Martian rover. 
\item SMAP, collected by NASA, offers soil moisture data acquired through spacecraft monitoring systems. 
\item PSM is sourced from eBay’s server machines and records metrics associated with their operational performance. 
\item SWaT includes sensor data from a water treatment infrastructure that operates continuously. 
\item NeurIPS-TS is a dataset introduced by~\cite{lai2021revisiting}, and GECCO and SWAN are its sub-datasets, which cover a variety of anomaly scenarios.
\item UCR consists of 250 sub-datasets, with each containing one-dimensional data that has a single anomaly segment.
\end{enumerate}

The statistical details of all the aforementioned datasets are summarized in Table~\ref{tab:table5}.
\input{Tables/table5}

\subsection{Evaluation Metrics}
\label{sec:Evaluation Metrics}
Time series anomaly detection involves identifying point-based anomalies (individual outliers) and range-based anomalies (continuous outlier segments). Traditional metrics suffer from limitations: threshold-based metrics (e.g., Precision, F1-score) rely on manual threshold tuning; threshold-free metrics (e.g., AUC-ROC, AUC-PR) fail to adapt to range-based anomalies and label misalignments; range-extended metrics (e.g., Range-AUC) still depend on buffer length parameters. Based on recent research~\cite{paparrizos2022volume,boniol2025vus}, the following focuses on the Volume Under the Surface (VUS) series, parameter-free and threshold-free metrics that address these limitations.

\subsubsection{Preliminaries of Key Metrics}
\label{sec:Preliminaries of Key Metrics}
\begin{itemize}
\item Threshold-based metrics: Discretize anomaly scores via a threshold to compute confusion matrix-based values (e.g., Precision, Recall, F1-score), but are sensitive to threshold choice.
\item Threshold-free metrics: Compute AUC-ROC/AUC-PR by iterating over all thresholds, but only adapt to point-based anomalies.
\item Range-AUC series: Extend labels with buffer regions to handle range anomalies, but require tuning the buffer length parameter \( \ell \).
\end{itemize}

\subsubsection{Parameter-Free Threshold-Free Metrics (VUS Series)}
\label{sec:Parameter-Free Threshold-Free Metrics (VUS Series)}
VUS upgrades ``area under the curve'' to ``volume under the 3D surface'' by iterating over all buffer lengths and thresholds, achieving fully parameter-free and robust evaluation for time series anomaly detection.

\textbf{Core Foundation: Continuous Label Extension. }
To handle range anomalies and label misalignments, extend discrete ground-truth labels \( label \in \{0,1\}^n \) (0=normal, 1=anomaly) to continuous labels \( label_\ell \) ( \( \ell \) = buffer length, default: half the time-series period). For an anomaly interval \( [s,e] \):
\begin{align}
label_{\ell i} = \begin{cases} 
\left(1 - \frac{|s - i|}{\ell}\right)^{\frac{1}{2}}, & s - \frac{\ell}{2} \leq i < s \land pred_i=1 \\
1, & s \leq i < e \\
\left(1 - \frac{|e - i|}{\ell}\right)^{\frac{1}{2}}, & e \leq i < e + \frac{\ell}{2} \land pred_i=1 \\
0, & \text{otherwise} 
\end{cases}
\end{align}
where \( pred_i \) is the binary prediction derived from anomaly score \( S_{Ti} \) and a threshold. If no predicted anomaly exists in the buffer region, \( label_{\ell i}=0 \); overlapping buffers take the maximum value. The components of the extended confusion matrix are defined as follows: the true positive count for class \(\ell\) is denoted as \(TP_\ell = label_\ell^\top \cdot pred\); the false positive count for class \(\ell\) is expressed as \(FP_\ell = (I - label_\ell)^\top \cdot pred\); the extended positive sample count for class \(\ell\) is given by \(P_\ell = \frac{(label + label_\ell)^\top \cdot I}{2}\), a formulation designed to avoid label distortion; and the range-adapted true positive rate for class \(\ell\) is computed as \(TPR_\ell = \frac{TP_\ell}{P_\ell} \cdot \sum_{R_i \in R} \frac{ER(R_i, P)}{|R|}\).

\textbf{VUS-ROC Calculation. }
Iterate over buffer length set \( L = [\ell_0, \ell_1,..., \ell_L] \) ( \( 0=\ell_0 < \ell_1 <...< \ell_L=\ell_{max} \) ) and threshold set \( Th = [Th_0, Th_1,..., Th_N] \) ( \( 0=Th_0 < Th_1 <...< Th_N=1 \) ). The volume under the \( TPR_\ell\text{-}FPR_\ell\text{-}\ell \) surface is:
\begin{align}
\text{VUS-ROC} = \frac{1}{4} \sum_{w=1}^L \sum_{k=1}^N \Delta^{(k,w)} \cdot |\ell_w - \ell_{w-1}| 
\end{align}
\begin{align}
\begin{split}
\Delta^{(k,w)} =& \underbrace{[TPR_{\ell_w}(Th_{k-1}) + TPR_{\ell_w}(Th_k)] \cdot [FPR_{\ell_w}(Th_k) - FPR_{\ell_w}(Th_{k-1})]}_{\text{Contribution of buffer } \ell_w} \\
&+ \underbrace{[TPR_{\ell_{w-1}}(Th_{k-1}) + TPR_{\ell_{w-1}}(Th_k)] \cdot [FPR_{\ell_{w-1}}(Th_k) - FPR_{\ell_{w-1}}(Th_{k-1})]}_{\text{Contribution of buffer } \ell_{w-1}}
\end{split}
\end{align}

\textbf{VUS-PR Calculation. }
The volume under the \( \text{Precision}_\ell\text{-}\text{Recall}_\ell\text{-}\ell \) surface is:
\begin{align}
\text{VUS-PR} = \frac{1}{2} \sum_{w=1}^L \sum_{k=1}^N \Delta^{(k,w)} \cdot |\ell_w - \ell_{w-1}|
\end{align}
\begin{align}
\begin{split}
\Delta^{(k,w)} =& \underbrace{\text{Precision}_{\ell_w}(Th_k) \cdot [\text{Recall}_{\ell_w}(Th_k) - \text{Recall}_{\ell_w}(Th_{k-1})]}_{\text{Contribution of buffer } \ell_w} \\
&+ \underbrace{\text{Precision}_{\ell_{w-1}}(Th_k) \cdot [\text{Recall}_{\ell_{w-1}}(Th_k) - \text{Recall}_{\ell_{w-1}}(Th_{k-1})]}_{\text{Contribution of buffer } \ell_{w-1}}
\end{split}
\end{align}

\subsection{Setting}
\label{sec:Setting}
In our experiments, we implemente SCAN using PyTorch, and all experiments are conducted on an NVIDIA GeForce RTX 3090 24GB GPU. The optimization is performed using the Adam optimizer with an initial learning rate of $10^{-3}$. We set the batch size as 32, the dimension of hidden states as 256, and the multi-scale as \{25, 5, 1\}. We employ a sliding window of length 2500 for time series processing and perform anomaly detection based on non-overlapping windows. To ensure a fair performance comparison, we refrain from applying the drop last trick in the inference phase~\cite{qiu2024tfb}. After obtaining the anomaly scores, we use the widely used SPOT~\cite{siffer2017anomaly} method to determine the threshold.

\section{Additional Experiments and Analysis}
\label{sec:Additional Experiments and Analysis}

\subsection{Multi-metrics Results}
\label{sec:Multi-metrics Results}
For a more comprehensive comparison, we compare our SCAN model with four recognized advanced methods across multiple metrics, including Accuracy (Acc), Affiliation-F1 (Aff-F1), AUC-ROC (A-R), AUC-PR (A-P), Range-AUC-ROC (R-A-R), Range-AUC-PR (R-A-P), VUS-ROC (V-R) and VUS-PR (V-P), as shown in Table~\ref{tab:table2}. SCAN achieves superior or comparable performance over all metrics, further validating its effectiveness.
\input{Tables/table2}

\subsection{UCR Benchmark}
\label{sec:UCR Benchmark}
\begin{figure}[h]  
    \centerline{\includegraphics[width=0.9\linewidth]{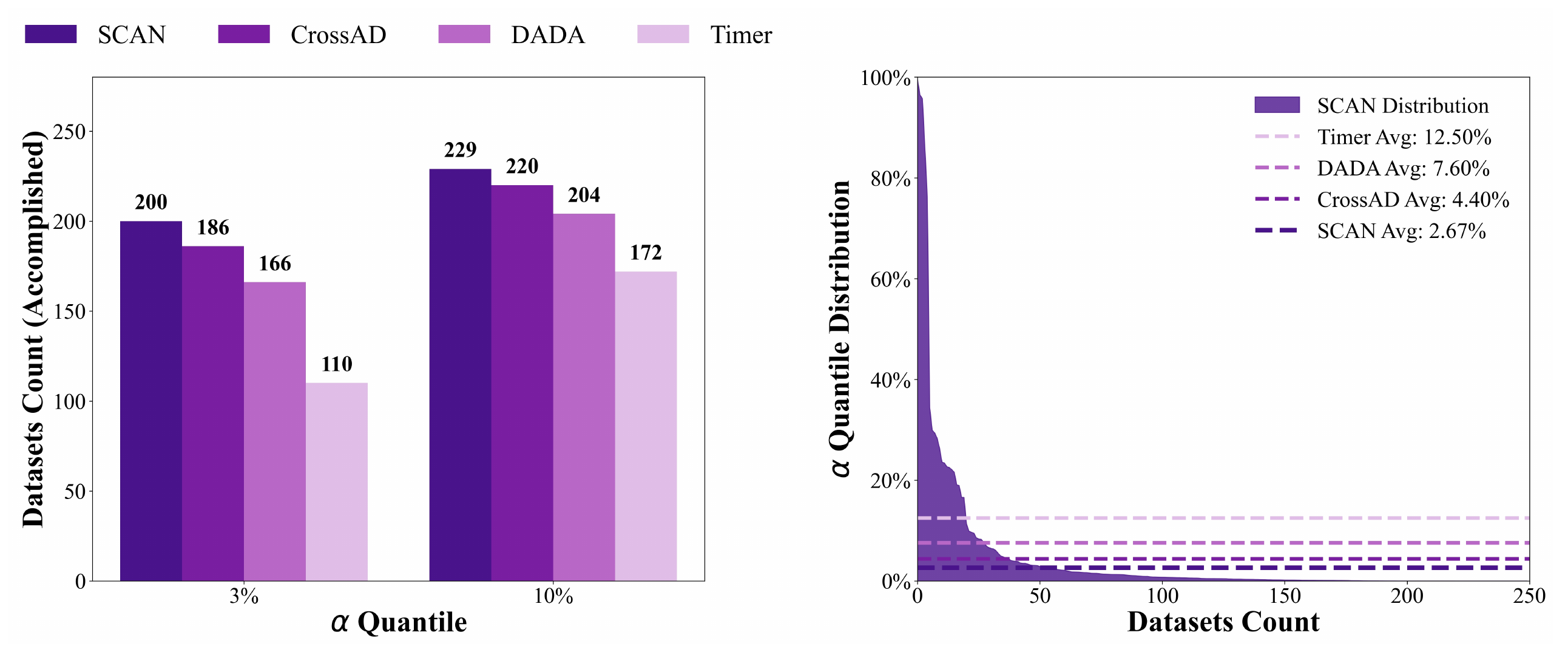}}
    \caption{Anomaly detection results based on the UCR benchmark. The left figure shows the number of datasets in which the model successfully detects anomalies at the 3\% and 10\% quantile levels, with higher counts indicating stronger detection capability. The right figure displays the quantile distribution of SCAN and the average quantiles of all models across all UCR dataset, with lower average quantiles indicating stronger detection capability.}
    \vspace{-10pt}
    \label{fig:appendix_fig1}
\end{figure}

\begin{figure}[p]  
\centering
\includegraphics[width=1\linewidth]{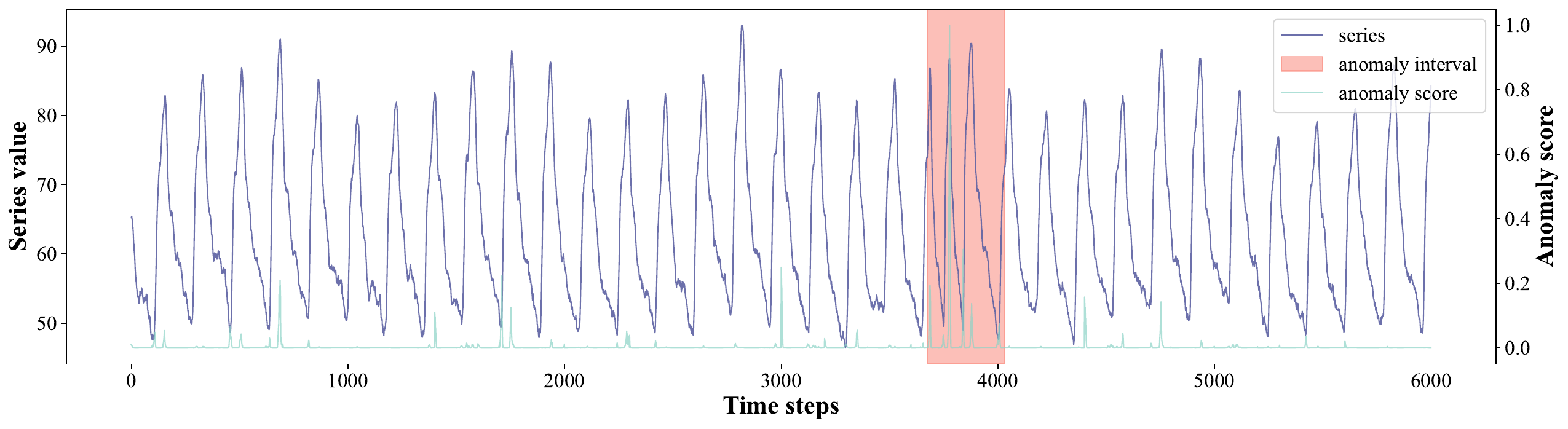}
\vspace{-10pt}
\caption{Visualization of 032\_UCR\_Anomaly\_DISTORTEDInternalBleeding4\_1000\_4675\_5033.txt.}
\label{fig:appendix_fig2}

\vspace{30pt}  

\includegraphics[width=1\linewidth]{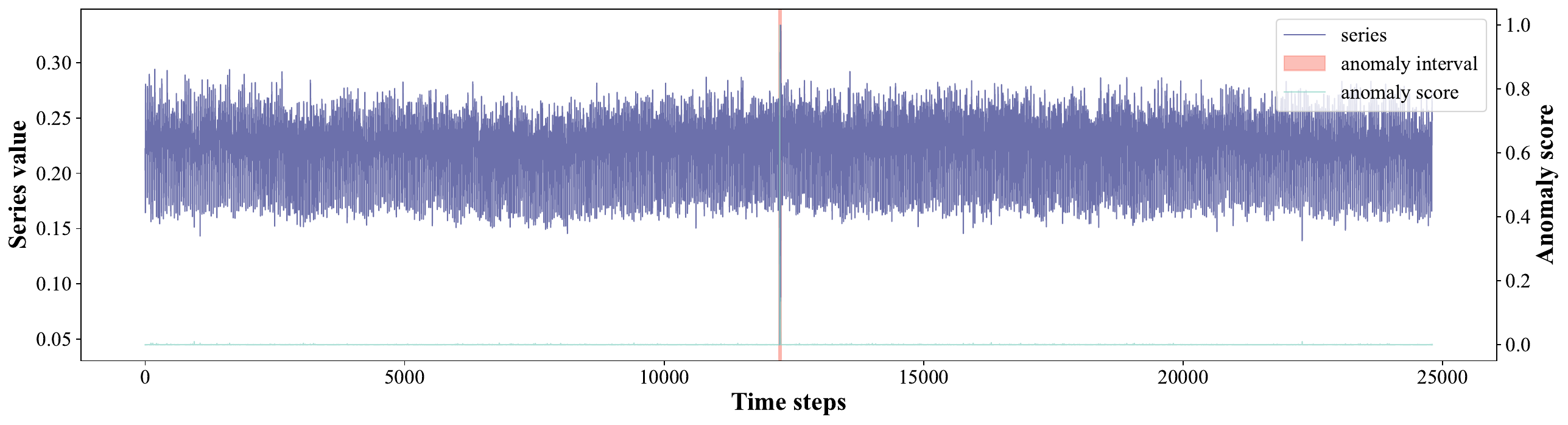}
\vspace{-10pt}
\caption{Visualization of 037\_UCR\_Anomaly\_DISTORTEDLab2Cmac011215EPG1\_5000\_17210\_17260.txt.}
\label{fig:appendix_fig3}

\vspace{30pt}

\includegraphics[width=1\linewidth]{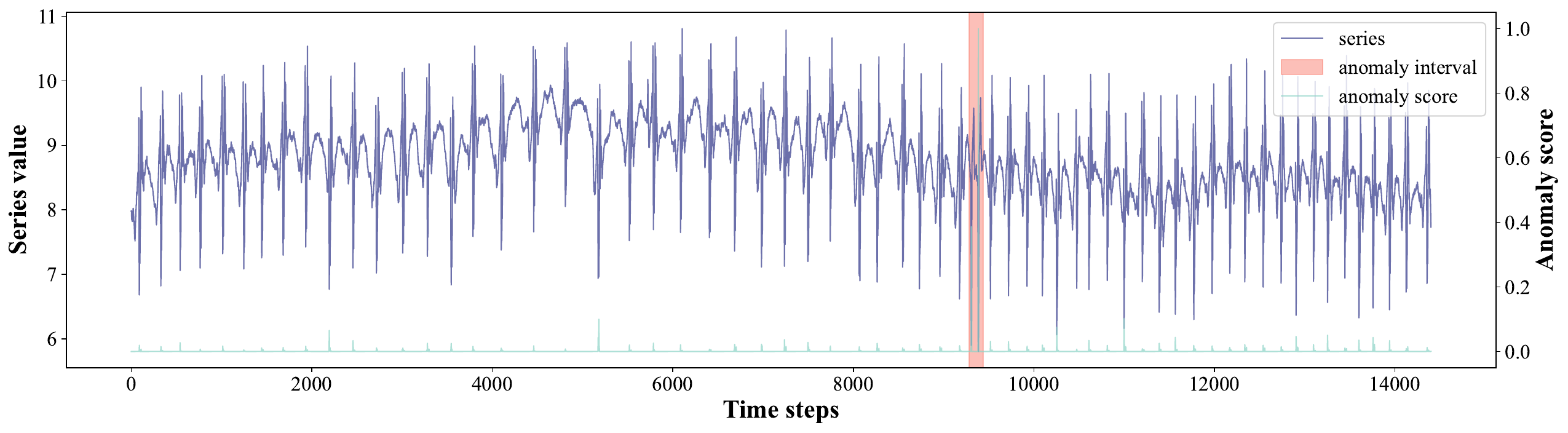}
\vspace{-10pt}
\caption{Visualization of 043\_UCR\_Anomaly\_DISTORTEDMesoplodonDensirostris\_10000\_19280\_19440.txt.}
\label{fig:appendix_fig4}

\vspace{30pt}

\includegraphics[width=1\linewidth]{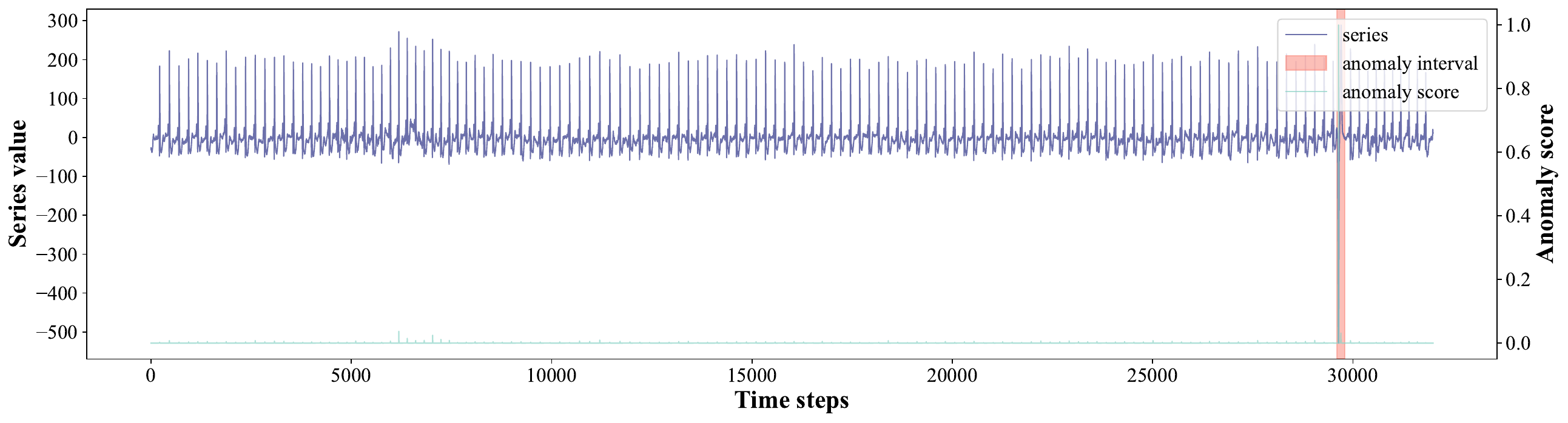}
\vspace{-10pt}
\caption{Visualization of 071\_UCR\_Anomaly\_DISTORTEDltstdbs30791AS\_23000\_52600\_52800.txt.}
\label{fig:appendix_fig5}
\end{figure}

\begin{figure}[p]
\centering
\includegraphics[width=1\linewidth]{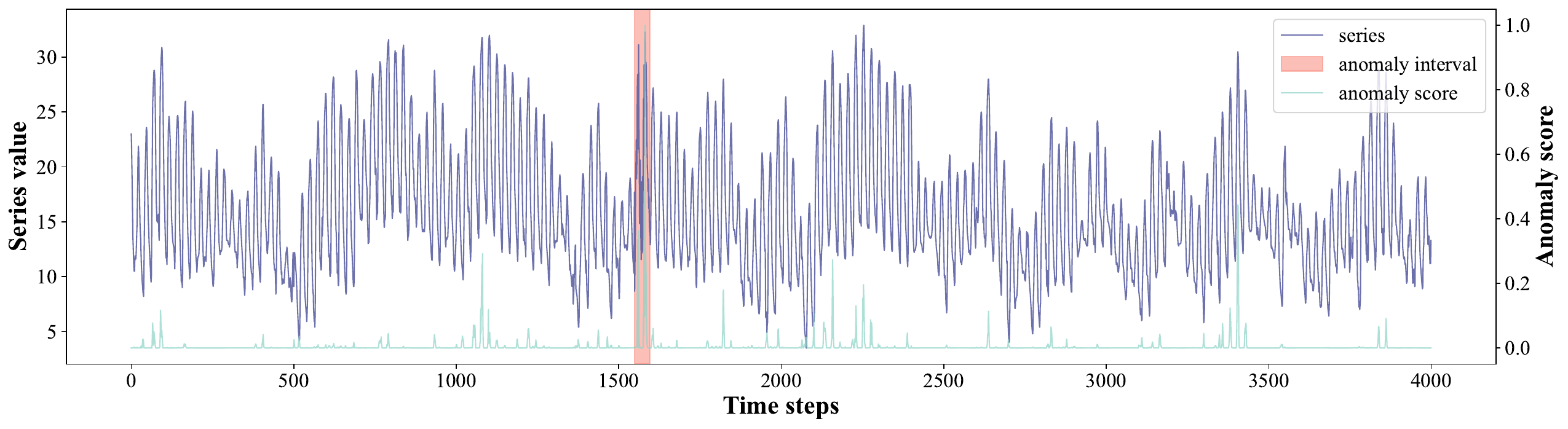}
\vspace{-10pt}
\caption{Visualization of 116\_UCR\_Anomaly\_CIMIS44AirTemperature4\_4000\_5549\_5597.txt.}
\label{fig:appendix_fig6}

\vspace{30pt}

\includegraphics[width=1\linewidth]{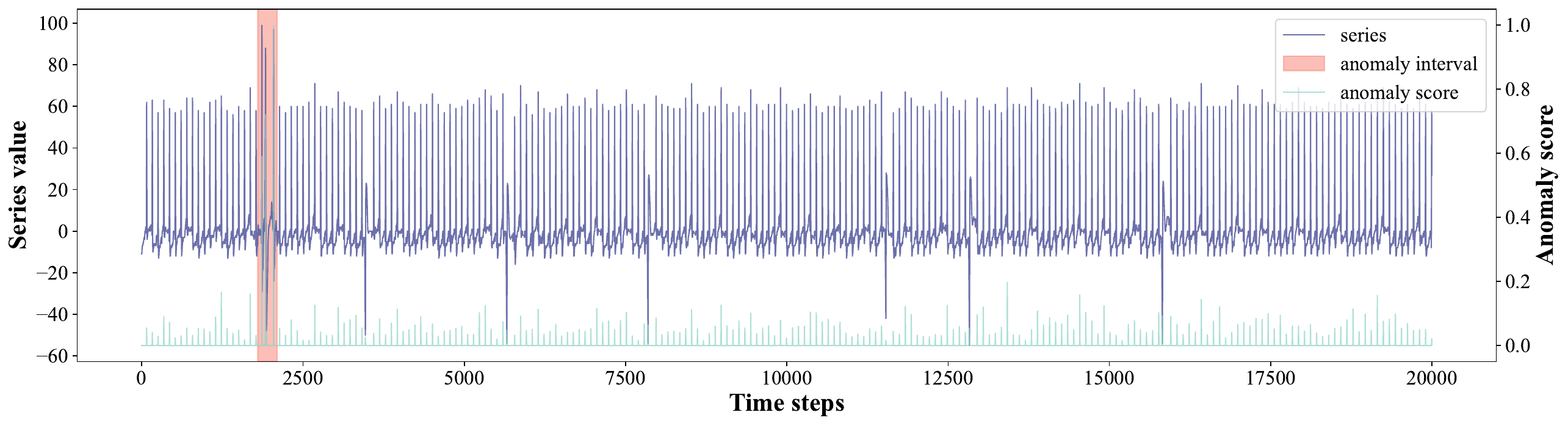}
\vspace{-10pt}
\caption{Visualization of 119\_UCR\_Anomaly\_ECG1\_10000\_11800\_12100.txt.}
\label{fig:appendix_fig7}

\vspace{30pt}

\includegraphics[width=1\linewidth]{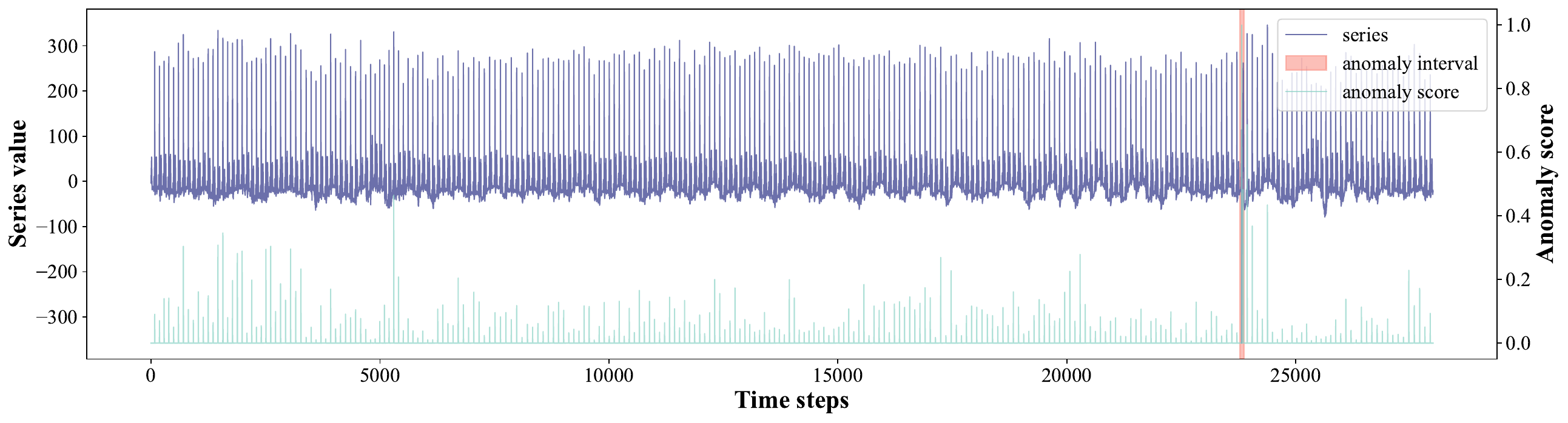}
\vspace{-10pt}
\caption{Visualization of 192\_UCR\_Anomaly\_s20101mML2\_12000\_35774\_35874.txt.}
\label{fig:appendix_fig8}

\vspace{30pt}

\includegraphics[width=1\linewidth]{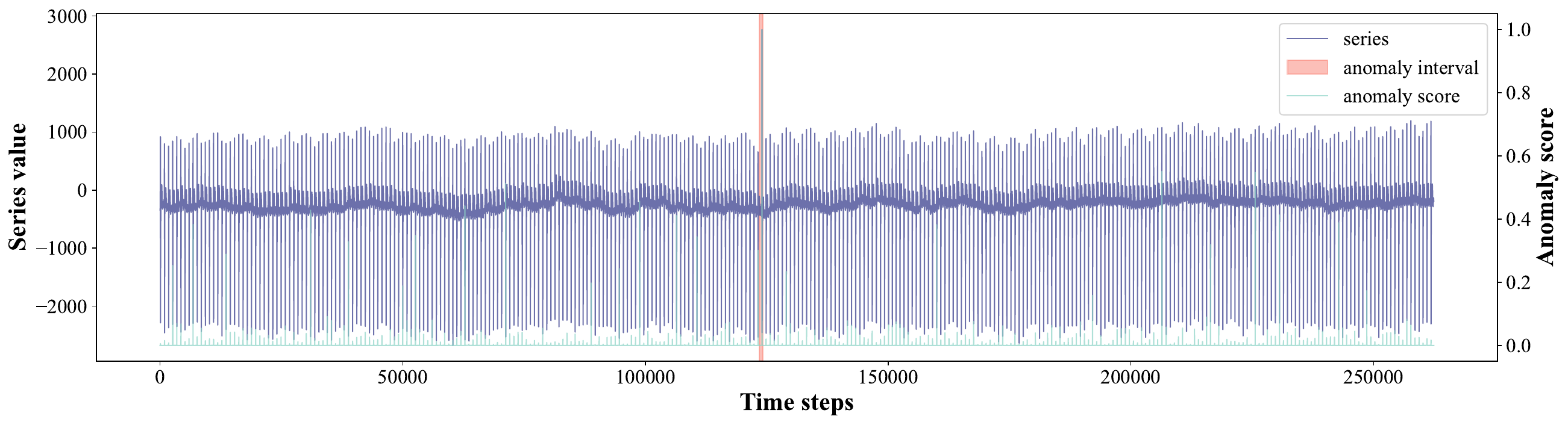}
\vspace{-10pt}
\caption{Visualization of 216\_UCR\_Anomaly\_STAFFIIIDatabase\_37216\_160720\_161370.txt.}
\label{fig:appendix_fig9}
\end{figure}

The UCR dataset~\cite{wu2021current} consists of 250 sub-datasets. Each sub-dataset contains a univariate time series with only a single anomaly segment. For the UCR dataset, the time series anomaly detection task aims to identify the location of the anomaly within the test set of each sub-dataset. Therefore, we compute the anomaly score for each time step in the test set and then rank them in descending order. Following the approach of Timer~\cite{liutimer}, anomaly detection is accomplished if the time step with the alpha quantile hits the labeled anomaly interval in the test set. We evaluate SCAN on the UCR dataset, comparing it against Timer~\cite{liutimer}, DADA~\cite{shentu2025towards}, and CrossAD~\cite{li2025crossad}, as shown in Figure ~\ref{fig:appendix_fig1}. The left figure shows the number of datasets in which the model successfully detects anomalies at the 3\% and 10\% quantile levels. The right figure displays the quantile distribution and the average quantile across all UCR datasets. SCAN achieves a higher number of successful detections and a lower average quantile, demonstrating its superior anomaly detection capability. Furthermore, we visualize the detection results of SCAN to highlight its robust anomaly detection capability, as shown in Figure~\ref{fig:appendix_fig2} to Figure~\ref{fig:appendix_fig9}.

\subsection{Complexity Analysis}
\label{sec:Complexity Analysis}
Let $j$ be the number of scales, $N_i$ be the number of patches at the $i$-th scale, $d_{\text{emb}}$ be the embedding dimension ($d_{\text{emb}} = d \times C$), $d_r$ be the clustering space dimension, $K$ be the number of clusters, and $L$ be the patch size. The computational cost of SCAN is the sum of the complexities of its constituent modules across $j$ scales and dominated by the following core parts: 

\begin{itemize}
\item Representation extraction: Constructing the similarity adjacency matrix $\mathbf{A}^{\text{sim}}_i$ involves computing all-pairs cosine similarities, which takes $O(N_i^2 d_{\text{emb}})$. The temporal-dependent adjacency matrix $\mathbf{A}^{\text{tim}}_i$ is constructed in $O(N_i^2)$. Neighborhood aggregation and centering operations involve matrix multiplications of size $(N_i \times N_i)$ and $(N_i \times d_{\text{emb}})$, resulting in a complexity of $O(N_i^2 d_{\text{emb}})$.
\item Clustering: Projecting patches into the clustering space takes $O(N_i d_{\text{emb}} d_r)$. The computation of the soft assignment $\mathbf{Y}_i^{\text{soft}}$ requires calculating distances between $N_i$ patches and $K$ centroids, taking $O(N_i K d_r)$. Constructing the same-cluster probability matrix $\mathbf{S}_i$ requires $O(N_i^2 K)$. The clustering loss $\mathcal{L}_{\text{clu}}$ involves matrix operations with $\mathbf{S}_i$ and the membership matrix $\mathbf{M}_i$, which scale as $O(N_i^2 K)$.
\item Fusion \& Reconstruction: This module first generates cluster-weighted representations by aggregating $K$ cluster centers according to the membership probabilities, which takes $O(N_i K d_r)$. The fused representations are then mapped back to the original input space via a linear head. Since each of the $N_i$ patches is projected to a patch of length $L$, the mapping cost is $O(N_i L d_{emb})$.
\end{itemize}

Thus, the overall time complexity is:
\begin{align}
\mathcal{O}\left( \sum_{i=1}^{j} \Big( N_i^2 (d_{\text{emb}} + K) + N_i (d_{\text{emb}}d_r + Kd_r + Ld_{emb}) \Big) \right)
\end{align}

\subsection{Parameter Sensitivity Analysis}
\label{sec:Parameter Sensitivity Analysis}
The primary hyperparameter affecting SCAN performance is the number of clusters. To analyze its impact on anomaly detection, we experimentally evaluate the performance of the model under different parameters. Effective clustering provides reliable support for anomaly detection. Table~\ref{tab:table6} shows the performance of the model at different cluster numbers $k$. It can be concluded that increasing the number of clusters generally improves model performance, with performance enhancement tending to become stable after a certain number of clusters is reached.
\input{Tables/table6}

\section{Broader impacts}
\label{sec:Broader impacts}
Time series anomaly detection plays a vital role in practical applications. Early detection of anomalies can prevent potential risks and reduce economic losses. This work addresses the trade-off dilemma between over-generalization and under-generalization of reconstruction-based methods, which helps promote the development of this field. Focusing on technical improvement, this work does not involve datasets that may contain sensitive information or privacy risks, and thus will not exert any negative impacts on society.

\newpage

%% file: Tables/table5.tex
\begin{table}[htbp]
  \centering
  \caption{Statistics of the datasets. The anomaly ratio denotes the abnormal proportion of the entire dataset.}
  \rowcolors{2}{white}{gray!10}
  \resizebox{0.9\linewidth}{!}{
    \begin{tabular}{ccccccc}
    \toprule
    \textbf{Dataset} & \textbf{Domain} & \textbf{Dimension} & \textbf{Training} & \textbf{Validation} & \textbf{Test(labeled)} & \textbf{Anomaly Ratio(\%)} \\
    \midrule
    MSL   & Spacecraft & 1     & 46,653 & 11,664 & 73,729 & 10.5 \\
    SMAP  & Spacecraft & 1     & 108,146 & 27,037 & 427,617 & 12.8 \\
    PSM   & Server Machine & 25    & 105,984 & 26,497 & 87,841 & 27.8 \\
    SMD   & Server Machine & 38    & 566,724 & 141,681 & 708,420 & 4.2 \\
    SWaT  & Water treatment & 31    & 396,000 & 99,000 & 449,919 & 12.1 \\
    GECCO & Water treatment & 9     & 55,408 & 13,852 & 69,261 & 1.25 \\
    SWAN  & Space Weather & 38    & 48,000 & 12,000 & 60,000 & 23.8 \\
    UCR   & Natural & 1     & 1,790,680 & 447,670 & 6,143,541 & 0.6 \\
    \bottomrule
    \end{tabular}}%
  \label{tab:table5}%
\end{table}%

%% file: Tables/table2.tex
\begin{table}[htbp]
  \centering
  \renewcommand{\arraystretch}{1.0}
  \caption{Multi-metrics results in the three real-world datasets. Higher values for all metrics indicate better performance. \textcolor[rgb]{ 1,  0,  0}{\textbf{Red}}: the best, \textcolor[rgb]{ .161,  .447,  .957}{\underline{Blue}}: the 2nd best.}
  \resizebox{0.9\linewidth}{!}{
  \setlength{\tabcolsep}{12pt}
    \begin{tabular}{c|c|cccccccc}
    \toprule
    Dataset & Method & Acc   & Aff-F1 & A-R   & A-P   & R-A-R & R-A-P & V-R   & V-P \\
    \midrule
    \multirow{4.5}[2]{*}{SMD} & \cellcolor{gray!10} ModernTCN & \cellcolor{gray!10} 0.9074  & \cellcolor{gray!10} 0.8316  & \cellcolor{gray!10} 0.7021  & \cellcolor{gray!10} 0.1401  & \cellcolor{gray!10} 0.7754  & \cellcolor{gray!10} 0.1628  & \cellcolor{gray!10} 0.7707  & \cellcolor{gray!10} 0.1596  \\
    & TimeMixer & 0.9091  & \textcolor[rgb]{ .161,  .447,  .957}{\underline{0.8408}} & 0.6795  & 0.1215  & 0.7549  & 0.1580  & 0.7711  & 0.1390  \\
    & \cellcolor{gray!10} TimesNet & \cellcolor{gray!10} 0.9066  & \cellcolor{gray!10} 0.8397  & \cellcolor{gray!10} \textcolor[rgb]{ .161,  .447,  .957}{\underline{0.7603}} & \cellcolor{gray!10} \textcolor[rgb]{ .161,  .447,  .957}{\underline{0.1627}} & \cellcolor{gray!10} \textcolor[rgb]{ .161,  .447,  .957}{\underline{0.8300}} & \cellcolor{gray!10} \textcolor[rgb]{ .161,  .447,  .957}{\underline{0.2320}} & \cellcolor{gray!10} \textcolor[rgb]{ .161,  .447,  .957}{\underline{0.8420}} & \cellcolor{gray!10} \textcolor[rgb]{ .161,  .447,  .957}{\underline{0.2040}} \\
    & KAN-AD & \textcolor[rgb]{ .161,  .447,  .957}{\underline{0.9161}}  & 0.8297  & 0.7406  & 0.1549  & 0.7682  & 0.1605  & 0.7657  & 0.1593  \\
    & \cellcolor{red!10} SCAN  & \cellcolor{red!10} \textcolor[rgb]{ 1,  0,  0}{\textbf{0.9202}} & \cellcolor{red!10} \textcolor[rgb]{ 1,  0,  0}{\textbf{0.8431}} & \cellcolor{red!10} \textcolor[rgb]{ 1,  0,  0}{\textbf{0.8163}} & \cellcolor{red!10} \textcolor[rgb]{ 1,  0,  0}{\textbf{0.2182}} & \cellcolor{red!10} \textcolor[rgb]{ 1,  0,  0}{\textbf{0.8736}} & \cellcolor{red!10} \textcolor[rgb]{ 1,  0,  0}{\textbf{0.3004}} & \cellcolor{red!10} \textcolor[rgb]{ 1,  0,  0}{\textbf{0.9081}} & \cellcolor{red!10} \textcolor[rgb]{ 1,  0,  0}{\textbf{0.3203}} \\
    \midrule
    \multirow{4.5}[2]{*}{PSM} & \cellcolor{gray!10} ModernTCN & \cellcolor{gray!10} 0.3016  & \cellcolor{gray!10} 0.7959  & \cellcolor{gray!10} 0.5846  & \cellcolor{gray!10} 0.3843  & \cellcolor{gray!10} 0.6550  & \cellcolor{gray!10} 0.4748  & \cellcolor{gray!10} \textcolor[rgb]{ .161,  .447,  .957}{\underline{0.6480}} & \cellcolor{gray!10} \textcolor[rgb]{ .161,  .447,  .957}{\underline{0.4689}} \\
    & TimeMixer & 0.2774  & 0.7499  & 0.5522  & 0.3447  & 0.6140  & 0.4089  & 0.5974  & 0.3807  \\
    & \cellcolor{gray!10} TimesNet & \cellcolor{gray!10} 0.2773  & \cellcolor{gray!10} 0.7970  & \cellcolor{gray!10} 0.5755  & \cellcolor{gray!10} 0.3701  & \cellcolor{gray!10} \textcolor[rgb]{ .161,  .447,  .957}{\underline{0.6617}} & \cellcolor{gray!10} \textcolor[rgb]{ .161,  .447,  .957}{\underline{0.4827}} & \cellcolor{gray!10} 0.6344  & \cellcolor{gray!10} 0.4373  \\
    & KAN-AD & \textcolor[rgb]{ 1,  0,  0}{\textbf{0.7347}} & \textcolor[rgb]{ 1,  0,  0}{\textbf{0.8534}} & \textcolor[rgb]{ .161,  .447,  .957}{\underline{0.6434}} & \textcolor[rgb]{ 1,  0,  0}{\textbf{0.4560}} & 0.6375  & 0.4514  & 0.6377  & 0.4523  \\
    & \cellcolor{red!10} SCAN  & \cellcolor{red!10} \textcolor[rgb]{ .161,  .447,  .957}{\underline{0.6298}}  & \cellcolor{red!10} \textcolor[rgb]{ .161,  .447,  .957}{\underline{0.8463}} & \cellcolor{red!10} \textcolor[rgb]{ 1,  0,  0}{\textbf{0.6452}} & \cellcolor{red!10} \textcolor[rgb]{ .161,  .447,  .957}{\underline{0.4478}} & \cellcolor{red!10} \textcolor[rgb]{ 1,  0,  0}{\textbf{0.7413}} & \cellcolor{red!10} \textcolor[rgb]{ 1,  0,  0}{\textbf{0.5895}} & \cellcolor{red!10} \textcolor[rgb]{ 1,  0,  0}{\textbf{0.7457}} & \cellcolor{red!10} \textcolor[rgb]{ 1,  0,  0}{\textbf{0.5851}} \\
    \midrule
    \multirow{4.5}[2]{*}{GECCO} & \cellcolor{gray!10} ModernTCN & \cellcolor{gray!10} \textcolor[rgb]{ .161,  .447,  .957}{\underline{0.9882}} & \cellcolor{gray!10} \textcolor[rgb]{ .161,  .447,  .957}{\underline{0.9018}} & \cellcolor{gray!10} 0.9595  & \cellcolor{gray!10} \textcolor[rgb]{ .161,  .447,  .957}{\underline{0.4325}} & \cellcolor{gray!10} 0.9733  & \cellcolor{gray!10} 0.5036  & \cellcolor{gray!10} 0.9694  & \cellcolor{gray!10} \textcolor[rgb]{ .161,  .447,  .957}{\underline{0.4819}} \\
    & TimeMixer & 0.9832  & 0.8989  & \textcolor[rgb]{ .161,  .447,  .957}{\underline{0.9620}} & 0.3559  & \textcolor[rgb]{ 1,  0,  0}{\textbf{0.9803}} & \textcolor[rgb]{ .161,  .447,  .957}{\underline{0.5106}} & \textcolor[rgb]{ .161,  .447,  .957}{\underline{0.9899}} & 0.4106  \\
    & \cellcolor{gray!10} TimesNet & \cellcolor{gray!10} 0.9860  & \cellcolor{gray!10} 0.8906  & \cellcolor{gray!10} 0.9173  & \cellcolor{gray!10} 0.3431  & \cellcolor{gray!10} 0.9327  & \cellcolor{gray!10} 0.3593  & \cellcolor{gray!10} 0.9834  & \cellcolor{gray!10} 0.4578  \\
    & KAN-AD & 0.8333  & 0.7904  & 0.8415  & 0.3259  & 0.8642  & 0.1755  & 0.8567  & 0.1732  \\
    & \cellcolor{red!10} SCAN  & \cellcolor{red!10} \textcolor[rgb]{ 1,  0,  0}{\textbf{0.9897}} & \cellcolor{red!10} \textcolor[rgb]{ 1,  0,  0}{\textbf{0.9591}} & \cellcolor{red!10} \textcolor[rgb]{ 1,  0,  0}{\textbf{0.9811}} & \cellcolor{red!10} \textcolor[rgb]{ 1,  0,  0}{\textbf{0.5491}} & \cellcolor{red!10} \textcolor[rgb]{ .161,  .447,  .957}{\underline{0.9748}} & \cellcolor{red!10} \textcolor[rgb]{ 1,  0,  0}{\textbf{0.5339}} & \cellcolor{red!10} \textcolor[rgb]{ 1,  0,  0}{\textbf{0.9957}} & \cellcolor{red!10} \textcolor[rgb]{ 1,  0,  0}{\textbf{0.6812}} \\
    \bottomrule
    \end{tabular}}%
  \label{tab:table2}%
\end{table}%

%% file: Tables/table6.tex
\begin{table}[htbp]
  \centering
  \caption{Parameter sensitivity analysis on cluster numbers $k$. Higher VUS-ROC (V-R) or VUS-PR (V-P) values indicate better performance. \textbf{Bold}: the best.}
  \rowcolors{2}{gray!10}{white}
  \resizebox{0.8\linewidth}{!}{
    \begin{tabular}{c|cc|cc|cc|cc|cc}
    \toprule
    Dataset & \multicolumn{2}{c|}{MSL} & \multicolumn{2}{c|}{SMAP} & \multicolumn{2}{c|}{PSM} & \multicolumn{2}{c|}{SWaT} & \multicolumn{2}{c}{GECCO} \\
    \midrule
    Metric & V-R   & V-P   & V-R   & V-P   & V-R   & V-P   & V-R   & V-P   & V-R   & V-P \\
    \midrule
    k=5   & 0.8092  & \textbf{0.3330 } & 0.5694  & 0.1385  & 0.7045  & 0.5323  & 0.5621  & 0.2409  & 0.9937  & 0.6271  \\
    k=10  & \textbf{0.8232 } & 0.3315  & \textbf{0.6114 } & \textbf{0.1578 } & 0.7176  & 0.5643  & 0.6811  & 0.3832  & \textbf{0.9958 } & 0.6785  \\
    k=15  & 0.8163  & 0.3137  & 0.6068  & 0.1556  & 0.7451  & 0.5846  & 0.6819  & 0.2851  & 0.9957  & 0.6736  \\
    k=20  & 0.8187  & 0.3194  & 0.6020  & 0.1519  & 0.7418  & 0.5821  & 0.7443  & 0.3881  & 0.9957  & 0.6795  \\
    k=25  & 0.8167  & 0.3151  & 0.6016  & 0.1516  & 0.7448  & 0.5843  & 0.7445  & 0.4107  & 0.9956  & 0.6757  \\
    k=30  & 0.8225  & 0.3191  & 0.6014  & 0.1521  & 0.7449  & 0.5835  & 0.7421  & 0.4881  & 0.9955  & 0.6698  \\
    k=35  & 0.8222  & 0.3224  & 0.6024  & 0.1521  & 0.7454  & 0.5844  & 0.7826  & 0.4697  & 0.9956  & 0.6737  \\
    k=40  & 0.8228  & 0.3207  & 0.6005  & 0.1514  & \textbf{0.7457 } & \textbf{0.5851 } & 0.7866  & \textbf{0.4986 } & 0.9957  & 0.6799  \\
    k=45  & 0.8230  & 0.3227  & 0.5996  & 0.1506  & 0.7449  & 0.5838  & 0.7713  & 0.4245  & 0.9956  & 0.6789  \\
    k=50  & 0.8222  & 0.3202  & 0.5991  & 0.1511  & 0.7452  & 0.5848  & \textbf{0.7891 } & 0.4395  & 0.9957  & \textbf{0.6812 } \\
    \bottomrule
    \end{tabular}}%
  \label{tab:table6}%
\end{table}%